\documentclass{ieeeaccess}
\usepackage{cite}
\usepackage{amsmath,amssymb,amsfonts}
\usepackage{algorithmic}
\usepackage[pdftex]{graphicx}
\usepackage{textcomp}
\usepackage{comment}
\usepackage{subcaption}
\usepackage{algorithm,algorithmic}

\usepackage{color}
\newcommand{\red}[1]{\textcolor{black}{#1}}

\def\BibTeX{{\rm B\kern-.05em{\sc i\kern-.025em b}\kern-.08em
    T\kern-.1667em\lower.7ex\hbox{E}\kern-.125emX}}
\begin{document}
\history{Date of publication, date of current version Oct 10, 2019.}
\doi{10.1109/ACCESS.2019.DOI}

\title{Redirection Controller Using Reinforcement Learning}
\author{\uppercase{Yuchen Chang}\authorrefmark{1},
\uppercase{Keigo Matsumoto}\authorrefmark{1}, \IEEEmembership{Student member, IEEE},
\uppercase{Takuji Narumi}\authorrefmark{1}, \IEEEmembership{Member, IEEE},
\uppercase{Tomohiro Tanikawa}\authorrefmark{1}, \IEEEmembership{Member, IEEE},
\uppercase{Michitaka Hirose}\authorrefmark{1}, \IEEEmembership{Member, IEEE}}
\address[1]{Graduate School of Information Science and Technology, The University of Tokyo, Bunkyō City 7-3-1, Tokyo 113-0033, JAPAN (e-mail: {chang, matsumoto, narumi, tani, hirose}@cyber.t.u-tokyo.ac.jp)}
\tfootnote{Y. Chang and K. Matsumoto contributed equally to this paper. Their names are in alphabetical order. \\
This work was supported in part by \red{JSPS KAKENHI Grants JP19H04149 and JP18J21379.}}

\markboth
{Y. Chang, K. Matsumoto, T. Narumi, T. Tanikawa, and M. Hirose: Redirection Controller Using Reinforcement Learning}
{Y. Chang, K. Matsumoto, T. Narumi, T. Tanikawa, and M. Hirose: Redirection Controller Using Reinforcement Learning}

\corresp{Corresponding author: Keigo Matsumoto (e-mail: matsumoto@cyber.t.u-tokyo.ac.jp).}

\begin{abstract}
There is a growing demand for redirected walking (RDW) techniques and \red{their application} to various virtual and real environments.
To apply appropriate RDW methods and manipulation according to the real and virtual environments, the \red{RDW controllers are predominantly used.}
\red{There are three types of RDW controllers:} direct scripted controller, generalized controller, and predictive controller.
The scripted controller \red{type} pre-scripts the mapping between the real and virtual environments.
The generalized controller \red{type} employs the RDW method and manipulation quantities according to a certain procedure depending on the user's position in relation to the real space. 
This approach has the potential to be reused in any environment; however, it is not fully optimized.
The predictive controller \red{type} predicts the user's future path using the user's behavior and manages RDW techniques.
This approach is highly anticipated to be \red{very} effective and versatile; however, it has not been sufficiently developed.
\red{This paper proposes} a novel RDW controller using reinforcement learning (RL) with advanced plannability/versatility.
Our simulation experiments \red{indicate} that the proposed method can reduce the number of reset manipulations, which is one of the indicators of the effectiveness of the RDW controller, compared to the generalized controller under real environments with many obstacles.
Meanwhile, \red{the experimental results also showed} that the gain output by the proposed method oscillates.
The results of \red{a} user study \red{conducted} showed that the proposed RDW controller can reduce the number of resets compared to the conventional generalized controller.
Furthermore, no adverse effects such as cybersickness associated with the oscillation of the output gain were \red{evinced}.
The simulation and user studies demonstrate that the proposed RDW controller with RL outperforms the existing generalized controllers and can be applied to users as it only causes cybersickness comparable to that in the case of conventional generalized controllers.
\end{abstract}


\titlepgskip=-15pt

\maketitle

\section{Introduction}
\label{sec:introduction}
\PARstart{W}{ith} the increasing popularity of virtual reality (VR) technologies, such as a head-mounted display (HMD), there is a growing trend \red{of} exhibitions where users can walk around freely in immersive virtual environments (IVEs).
To walk through a virtual environment (VE), many methods have been proposed and applied \red{thus} far.
One \red{such method} involves tracking the user's position and reflecting it onto the user's virtual position in a VE.
However, to replicate IVEs through such a system, an equally large real space is required.

To overcome this problem, redirected walking (RDW) or redirection was proposed. \red{This approach} compresses a large VE into a significantly smaller real room while maintaining the sense of real walking\cite{Razzaque2002}.
Numerous redirection techniques have been proposed \red{to date} such as rotation, translation, and curvature gains.
However, it is next to impossible to freely walk around a VE by using only these individual methods.

Therefore, {\it redirection controllers} that apply optimal redirection methods at the optimal place and timing have been proposed.
Redirection controllers are roughly classified into three types: {\it scripted}, {\it generalized}, and {\it predictive} \cite{Nilsson2018}.
{\it Scripted controllers} are used for mapping the tracked space to VR space, as determined by the developer in advance. Although this method is the most effective when a user walks along a predetermined route, it fails to deal with unexpected movements.
{\it Generalized controllers} guide a user's walking route to specific patterns in a \red{real space}.
Although both these methods are highly versatile and adaptive to various walking paths, the optimal method is known to differ depending on the spatial arrangement of VEs \cite{Hodgson2013, Hodgson2014}.
{\it Predictive controllers} are used for analyzing the tracked space and VE and predicting the walking path of the user to perform appropriate redirection operations at the appropriate time.

Considerable research has been conducted to analyze the spatial composition of a VE and plan effective redirection techniques \cite{Zmuda2013, Nescher2014}. However, these methods \red{are not adequately} developed for practical use.
In the conventional method, it is necessary for experts to adjust the optimal RDW controller and gain according to the actual environment and VR environment.
There is also a need for a method to deal with difficult situations such as \red{real spaces} with complex shapes or cases where obstacles exist in the \red{real space}.

Meanwhile, recent studies have \red{been exploring} the possibility of applying reinforcement learning (RL) to many fields, such as board games \cite{silver2018general}, robotics \cite{gu2017deep}, and autonomous cars \cite{sallab2017deep}, owing to the development of deep learning (DL) in recent years.
By using deep RL (DRL), it is possible to execute objectives without the need for procedural knowledge.
Based on these features of RL, it is considered that RDW, which conventionally has to be customized according to each real environment and VR environment by an expert, can be automatically applied.
Thus, it would be natural to attempt to use RL to control redirection.

Therefore, in this study, we attempted to construct a novel generalized redirection controller by using DRL.
It is expected that this novel controller will be able to cope with an environment with obstacles, without the need for individual adjustments.
The contributions of this work are as follows:
\begin{enumerate}
\item Proposal and implementation of a novel generalized redirection controller using RL
\item Evaluation of the conventional generalized controller and the RL-based redirection controller in a \red{real space} with obstacles
\item Evaluation of the conventional generalized controller and the RL-based redirection controller in different types of walking paths in a VE
\item Evaluation of the RL-based redirection controller via a user study
\end{enumerate}

The remainder of this paper is organized as follows.
In Section II, we review the literature related to redirection and RL.
Section III describes the proposed redirection controller using RL.
Section IV describes common algorithms and simulation framework of the experiments conducted in this study.
In Section V, we propose and evaluate a heuristic redirection controller for a \red{real space} with obstacles as a preliminary experiment.
In Section VI, we describe the simulation results of applying the RL-based controller to individual redirection manipulations.
Sections VII and VIII present the comparisons of the heuristic redirection controller with the redirection controller using RL in a \red{real space} with obstacles and in different types of VEs.
Section IX discusses the user study for our proposed techniques.
Section X presents generalized discussions of this work.
Finally, Section XI summarizes the contributions of this work and presents future enhancements.

\section{Related Work}
In this section, we introduce relevant research on the redirection method, control of redirection, and DRL.

\subsection{Approaches to Redirected Walking}
RDW is a methodology for walking a vast VE within a limited tracked space. It was first proposed by Razzaque \cite{Razzaque2002}.
Since then, many redirection techniques have been proposed.
A recent survey on redirection provides a comprehensive overview of works in this area \cite{Nilsson2018}.
Suma et al. \cite{Suma2012} attempted to classify redirection techniques from multiple viewpoints.
One of these classifications involves dividing these techniques into subtle or overt manipulations.
The advantage of subtle manipulations is that the user does not realize that she/he is being manipulated, thus maintaining a sense of immersion.
In contrast, overt manipulations are generally more effective; however, the user realizes the manipulation, and the immersion decreases.

Subtle manipulations include rotation, translation, curvature and bending gains \cite{Nilsson2018}.
Rotation gains can be used to change the amount of rotation of the tracked space and VE when the user rotates on the spot.
Translation gains can be used to change the amount of movement of the user in the real and VR spaces.
Curvature gains can shift the walking direction by gradually rotating the user during movement; this allows the mapping of a straight walking path in the VE to a circular path in the tracked space.
Bending gains can also change the walking direction of a user walking on a curved path in a tracked space by using a method resembling the curvature gains, and it allows for the mapping of a curved path in the VE to a different curvature in the tracked space.
The thresholds of these gains are verified through psycho-physical experiments.
Bruder et al. \cite{Bruder2009} evaluated the thresholds of rotation gains and found that, compared to the VE, users can be physically turned approximately 30.9\% more and 16.2\% less.
Similarly, Steinicke et al. \cite{Steinicke2010} found that users can physically turn approximately 49\% more and 20\% less compared to virtual rotation.
They also noted that imperceptible translation gains range from down-scaling by 14\% to up-scaling by 26\%. 
Regarding curvature gains, studies have suggested that users can be redirected in VEs along the circumference of circles with radii of 22 m \cite{Steinicke2010}, 11.6 m \cite{Grechkin2016}, or 6.4 m \cite{Grechkin2016}.
Regarding bending gains, Langbehn et al. \cite{Langbehn2017} reported that curves in the real world with radii of curvatures of $r_{real}= 2.5\ m$ and $r_{real}=1.25\ m$ can be mapped to ones with $r_{virtual}= 10.875\ m$ and $r_{virtual}=4.0625\ m$ in the VE, respectively.
In some gains, results differ depending on the experimental setup; this is assumed to be due to the differences in the attributes of participants, experimental procedures, VEs, and equipment used in experiments.
Additionally, investigations have reported that the threshold does not change even if the curvature and translation gains are simultaneously applied \cite{Grechkin2016}.

Although performing redirection with such subtle operations is ideal, it is unfortunately not realistic.
The user must turn to the center of the tracked space by performing the reset operation when reaching near the boundary of the tracked space.
Suma et al. \cite{Suma2012} classified these operations as overt methods because they interrupt the user's VE experience in some way.
Several methods have been proposed for redirecting users to the center of the tracked space, the most promising of which is the 2:1-Turn method introduced by Williams et al. \cite{Williams2007}.
In the 2:1-Turn method, a user is instructed to stop and physically turn, and the VE rotates at twice her/his speed.
Thus, the user physically turns $180^{\circ}$ and is rotated $360^{\circ}$ in the VE.
Through such an operation, the user can move forward again; however, the feeling of immersion may decrease because the user is forced to stop and rotate on the spot \cite{Peck2009}.
A resolution to this problem is to display distractors, such as a sphere or a butterfly moving around them that guides the user to complete the reset operation\cite{Peck2009}. However, selecting an appropriate distractor that matches the content in the VE may be difficult, and hence this method is not sufficiently versatile.
Therefore, these overt operations should be avoided as much as possible.

\subsection{Redirection Controller}
To walk in a vast VE freely, the methods introduced in the previous section must be combined at an optimum time.
Redirection controllers have been proposed to perform individual methods at the optimum time.
Nilsson et al. \cite{Nilsson2018} stated that such a redirection controllers can be roughly classified into three types\red{: scripted, generalized, and predictive.}

{\it Scripted controllers} assume that the user walks on a predetermined walking route.
By using redirection techniques in a predetermined place, it is possible to complete the VR experience by using only subtle operations.
However, while using scripted controllers, users experience difficulty when deviating from a predetermined course, and the redirection techniques must be changed each time the spatial configurations of the tracked space and VE change. 

In general, {\it generalized controllers} can be applied regardless of the spatial structure of the VE or tracked space, thus retaining high versatility.
Various methods such as Steer-to-Center, Steer-to-Multiple-Waypoint, and Steer-to-Orbit have been proposed as generalized controllers \cite{razzaque2005redirected}.
Steer-to-Center is a method to guide users to the center of a tracked space; Steer-to-Multiple-Waypoint is a method to guide users to multiple points; and Steer-to-Orbit is a method to guide users to a circular walking route whose center is also the center of the tracked space.
Some researchers have verified the best method for generalized controllers according to the spatial configuration of the VE; they showed that Steer-to-Center is effective in nearly all cases, whereas Steer-to-Orbit is effective with live users \cite{Hodgson2013, Hodgson2014}.
However, there is a limit to the optimization for specific space configurations, often requiring overt operations such as resetting.

{\it Predictive controllers} attempt to perform the optimal redirection operation at the optimal timing by predicting the walking path of the user based on the spatial configuration of the VE and tracked space.
Zmuda et al. \cite{Zmuda2013} proposed an algorithm called FORCE as a redirection controller optimized in a constrained environment.
Similarly, Nescher et al. \cite{Nescher2014} presented an algorithm called MPCRed for dynamically choosing appropriate redirection controllers to optimize space and minimize costs.
Methods have also been proposed for extracting a walkable route by using a graph or navigation mesh and predicting the walking route of the user to perform a predictive redirection control \cite{Zank2017, Azmandian2016automated}.
Moreover, Chen et al. \cite{Chen2018} applied redirection algorithms to irregularly shaped and dynamic \red{real environments}.
To improve predictive controller accuracy, Zank and Kunz \cite{Zank2017} conducted a \red{study in which they} predicted the user's route by generating a skeleton graph from the navigation mesh.
\red{In addition, a method called alignment-based redirection controller (ARC) guided the user so that their proximity to an obstacle in the physical environment matched their proximity to an obstacle in the virtual environment as much as possible, allowing the user to explore the vast and complex virtual environment while minimizing collisions with obstacles in the physical environment\cite{williams2021arc}.}
More recently, machine learning technologies have been applied to redirection \cite{Cha2018, lee2019real}.
By using long short-term memory (LSTM), which is an artificial neural network that can process time series data and make predictions based on them, Cha et al. \cite{Cha2018} predicted the user's future path and provided corresponding manipulations for redirection.
\red{Furthermore, Lee et al. \cite{lee2019real} proposed real-time optimal planning for redirection using deep Q-learning.}

Although various trials have been conducted using the redirection controller, it is difficult to select an optimal strategy when the real space comprises obstacles or has a complicated shape.
Therefore, we attempted to create a controller that autonomously constructs the optimal strategy when multiple obstacles exist in the real space by using RL.

\subsection{Deep Reinforcement Learning}
In this subsection, we briefly explain about DRL and the methods used in this study. 

\subsubsection{Reinforcement Learning}
\label{sec:RL}
RL is a \red{form} of machine learning, and is often used to automatically produce a series of actions, such as robot locomotion and autonomous driving\red{\cite{kaelbling1996reinforcement, sutton2018reinforcement, cascone2021dtpaal}}.
Instead of using identifying labels, RL uses positive and negative rewards to evaluate output actions, leading to current environmental results. The use of RL is advantageous in that people who use RL do not need to know the optimal actions required toward goal setting. Some common examples that could use RL as a decision provider are trash-finding robots, pole balancing, and GO playing \cite{sutton2018reinforcement}.

An RL can often be described as a Markov decision process with a 4-tuple (S, A, P, and R), where $S\in\mathbb{R}^n$ is the observable-state set, $A\in\mathbb{R}^m$ is the possible action set, $P\in [0, 1]$ is the probability for the state to transit from $s_t$ to $s_{t+1}$ owing to action $a_t$, and $R\in\mathbb{R}$ is the reward received after the transition of state. Policy $\pi_{\theta} : S\to A\in\mathbb{R}^m$ is the function that outputs the best action (i.e., the most reward-receivable action) with the current observation of the environment. $\theta$ in the policy $ \pi _{\theta} $ implies all parameters in the policy. The learning part of RL could be explained as follows:

\begin{itemize}
\item \red{The} agent observes the environment as state $s_t$

\item According to $s_t$ and learned policy $\pi _t$, the agent outputs action $a_t$

\item The environment changes \red{owing} to $a_t$, and the given reward $r_t$ is recompensed to the agent

\item ($r_t, s_t, a_t$) is recorded into the \red{episode} memory, and time moves on to $t+1$; all the aforementioned steps are then repeated until $ t_{max} $ is reached

\item \red{The} agent updates policy $ \pi $ to $ \pi* $ according to the \red{episode}
\end{itemize}

\subsubsection{Learning Methods}
\label{sec:learning}
There are two common RL approaches for resolving the given problem: (1) value-based and (2) policy-based. These two \red{approaches} differ with regard to what RL learns.

The value-based \red{approach} updates the value (i.e., estimated-reward) function as a reference for the policy to decide which action to take. Q learning \cite{watkins1992q} is a common value-based method \red{that} stores and updates \red{the} expected future reward, named the Q value, of every state-action tuple. The value function can be expressed as follows:
\begin{equation}
V^{\pi}(s) = \mathbb{E}\left[  \sum_{i=1}^T  \gamma ^{i-1}r_i | s_t = s\right]
\end{equation}

where $\gamma\in [0, 1]$ is the discount factor, which shows the importance of considering the future reward.

In contrast, the policy-based \red{approach} directly adjusts the policy parameter $\theta$. If the policy function $\pi(s, a, \theta)$ is differentiable with respect to $\theta$, we can use policy gradient methods \cite{sutton2000policy} to update $\theta$:

\begin{equation}
\begin{split}
\theta_{i+1} &= \theta_{i} + \alpha_i\frac{\partial R}{\partial \theta_i} \\
where \; \frac{\partial R}{\partial \theta_i} &= \sum_{s}Pr^\pi(s) \sum_{a}\frac{\partial \pi(s, a)}{\partial \theta} Q^{\pi}(s, a)
\end{split}
\end{equation}

The policy-based \red{approach} could perform better for a large state or action space (for example, rational number space), as value-based methods must remember all state-action pairs. A well-known \red{method} of the policy-based \red{approach} is REINFORCE \cite{williams1992REINFORCE}.

A method that combines these two \red{approaches} is the Actor--Critic method \cite{konda2000actor}. It consists of two elements: the Actor, which is a policy that controls the output actions, and the Critic, which is a value function that evaluates the expected return and is used to update the actor. In this study, we used a variation of the Actor--Critic method as our RL architecture.

\subsubsection{Deep Reinforcement Learning and Proximal Policy Optimization Algorithm}

DL has been receiving increasing attention owing to its satisfactory results in several research fields. It can create an approximation of complex nonlinear functions by learning from sufficiently large datasets; this characteristic appropriately matches that of RL, as RL is often used in continuous serial action-deciding problems, such as robot feedback control. In particular, DRL is simply ordinary RL with the learning target (i.e., value function or policy) replaced by deep neural networks.

In this study, a well-known DRL architecture called the asynchronous advantage actor--critic (A3C) method proposed by \cite{mnih2016asynchronous} is used. In the following discussion, we explain the A3C method by dividing it in into three parts.
\begin{itemize}
\item \textbf{Asynchronous}

Multiple agents work independently in separate environments. After $ t_{max} $ time-steps or episodes, separate agents will have their own \red{episode} to update the core neural network. The asynchronous feature speeds up DRL, especially when working on a CPU.

\item \textbf{Advantage term}

The Advantage term is \red{typically} used to update the policy network. It shows the amount of benefits that would be obtained when taking actions. In this study, we considered the estimator used by Mnih et al. \cite{mnih2016asynchronous}:
\begin{equation}
\hat{A_t} = \left( (\sum_{i=t}^{T-1} \gamma ^{i-t} r_i) + \gamma ^{T-t} V(s_T)\right) - V(s_t),
\label{eq:advantage}
\end{equation}
where the first term on the right-hand side represents the estimated value after $T$ time-steps, and the second term represents the current value. 
The generalized advantage estimation (GAE)\cite{schulman2015high}, which is a framework that allows a single equation to generally express how far ahead steps are \red{considered} when calculating compensation in consideration of advantage, proposed by  could be written as

\begin{eqnarray}
\label{eq:gae}
\hat{A_t} & = & \sum_{i=t}^{T-1} (\gamma\lambda) ^{i-t} \delta _i \\
where\ \delta _i & = & r_t + \gamma V(s_{t+1}) - V(s_t)
\end{eqnarray}

where $\lambda \in [0,1]$ is an exponentially weighted hyperparameter. Equation (\ref{eq:advantage}) would be equal to (\ref{eq:gae}) if $\lambda = 1$, and $\delta _i$ is a one-step advantage estimator.

\item \textbf{Actor--Critic}

The same explanation as that in Section \ref{sec:learning} holds for the Actor--Critic part, with one or two deep neural networks receiving state parameters and outputting actions and estimated values.
\end{itemize}

\subsubsection{Method Employed}
To update the neural network part of DRL, a policy gradient optimizer is often used to simulate the policy gradient to perform gradient ascent. Based on the given loss function, the optimizer can evaluate how good or bad the current output is, and attempt to update the parameters of the neural network to minimize (or maximize) the loss function. In this study, we used a particular loss function for the policy gradient method called the proximal policy optimization (PPO), which was proposed by \cite{Schulman2017}. PPO updates the neural network using the following policy loss function (or objective function):
\begin{equation}
\begin{split}
L_t^{CLIP} (\theta) \\ &=\hat{\mathbb{E}_t}\left[min\left(r_t(\theta)\hat{A_t}, clip(r_t(\theta), 1-\epsilon, 1+\epsilon)\hat{A_t}\right)\right]
\end{split}
\end{equation}
where $ r_t(\theta) = \frac{\pi_{\theta}(a_t|s_t)}{\pi_{\theta_{old}}(a_t|s_t)}, \epsilon $ is the coefficient that normally ranges \red{between} 0.1 and 0.3, and $A_t$ is the advantage function. In addition, when the update step is too large, the clipping part $ clip(r_t(\theta), 1-\epsilon, 1+\epsilon) $ slows down the learning process down in order to increase stability.

As we used the A3C method, wherein the value function and policy both use a neural network, we considered the value function when updating the neural network by using the aforementioned loss. The actual loss function can be expressed as follows:

\begin{equation}
\begin{split}
L_t^{CLIP+V F+S}(\theta) \\
=\hat{\mathbb{E}_t}\left[L_t^{CLIP}(\theta) - c_1L_t^{V F}(\theta) + c_2S[\pi_{\theta}](s_t)\right]
\end{split}
\end{equation}

where $c_1$ and $c_2$ are coefficients, $L_t^{VF}$ is the residual sum of squares of the actor output and the target value function calculated from the current reward, and $S$ is the entropy bonus for encouraging DRL to explore.

An optimizer is also essential to update the neural network. It aims to find a correct way to update the parameters in order to minimize or maximize the loss function. In this study, we used the Adam optimizer \cite{kingma2014adam}, which employs the gradient of the loss function to update the neural network. Adam introduces the first and second moments of the gradient to maintain a regular/constant step size in sparse or dense gradient area. The update process of Adam can be expressed as follows:

\begin{align}
    \label{eqn:adam-1}
     &m_{t+1} = \beta_1 m_t+(1-\beta_1) \nabla L(\theta_t) \\
     &v_{t+1} = \beta_2 v_t+(1-\beta_2) \nabla L(\theta_t) ^2 \\
     &\hat{m}=\frac{m_{t+1}}{1-\beta_1^t} , 
    \hat{v}=\frac{v_{t+1}}{1-\beta_2^t} \nonumber \\
     &\theta_{t+1} = \theta_t - \alpha\frac{\hat{m}}{\sqrt{\hat{v}}+\epsilon}
    \label{eqn:adam-2}
\end{align}

where $\beta_1$, $\beta_2$ and $\epsilon$ are coefficients, and $m$ and $v$ are the exponential moving average of the first and second moment, respectively. 
In the case of a sparse or dense gradient area of the loss function, the $\sqrt{v}$ term in the denominator of (\ref{eqn:adam-2}) helps in maintaining the step size of every update as $\theta$. 

\section{Redirection Controller using RL}

\subsection{Concept}
The purpose of this novel redirection controller is to perform optimized redirection techniques for arbitrary VEs and tracked spaces as well as multiple users online.
Therefore, the RL is applied to learn the redirection controller, with the aim of minimizing negative rewards corresponding to the side effects of each redirection technique.
In this study, RL was considered because sufficient training data on redirection does not exist. In addition, learning in RL is performed without any specific measures and by providing only rewards.

We selected curvature, translation, and rotation gains during resetting as objects to be manipulated.
As the input of RL, we provided the user's position, orientation, and distance to the surrounding obstacles (boundaries and other users) in the tracked space.
When the target was known, we also provided the distance to the target in the VE as input.

\subsection{Overview of Redirection Controller} 
We attempted to adjust the values of translation, rotation, and curvature gains through RL.
Based on previous research on detection thresholds for RDW \cite{Steinicke2010, Grechkin2016}, the translation gains were limited to the range of 0.86--1.26 and the curvature gains were limited to \red{the range of $-$0.133--0.133 $[m^{-1}]$}.
The rotation angle during the reset operation can take an arbitrary value from $0^\circ$ to $360^\circ$.
Generally, in the reset operation, the user stops on the spot and rotation gains are applied while the user rotates $360^\circ$ in the VE, thereby changing the user-traveling direction in the \red{real space}.
The rotation angles applied in this study can be realized by setting an appropriate rotation gain from 0.5 to 1.
The output is a vector normalized from $-$1 to 1.
This vector is converted to translation gains, rotation angles at reset, and curvature gains by multiplying appropriate coefficients.

\begin{figure}[tb]
 \centering
 \includegraphics[width=0.8\columnwidth]{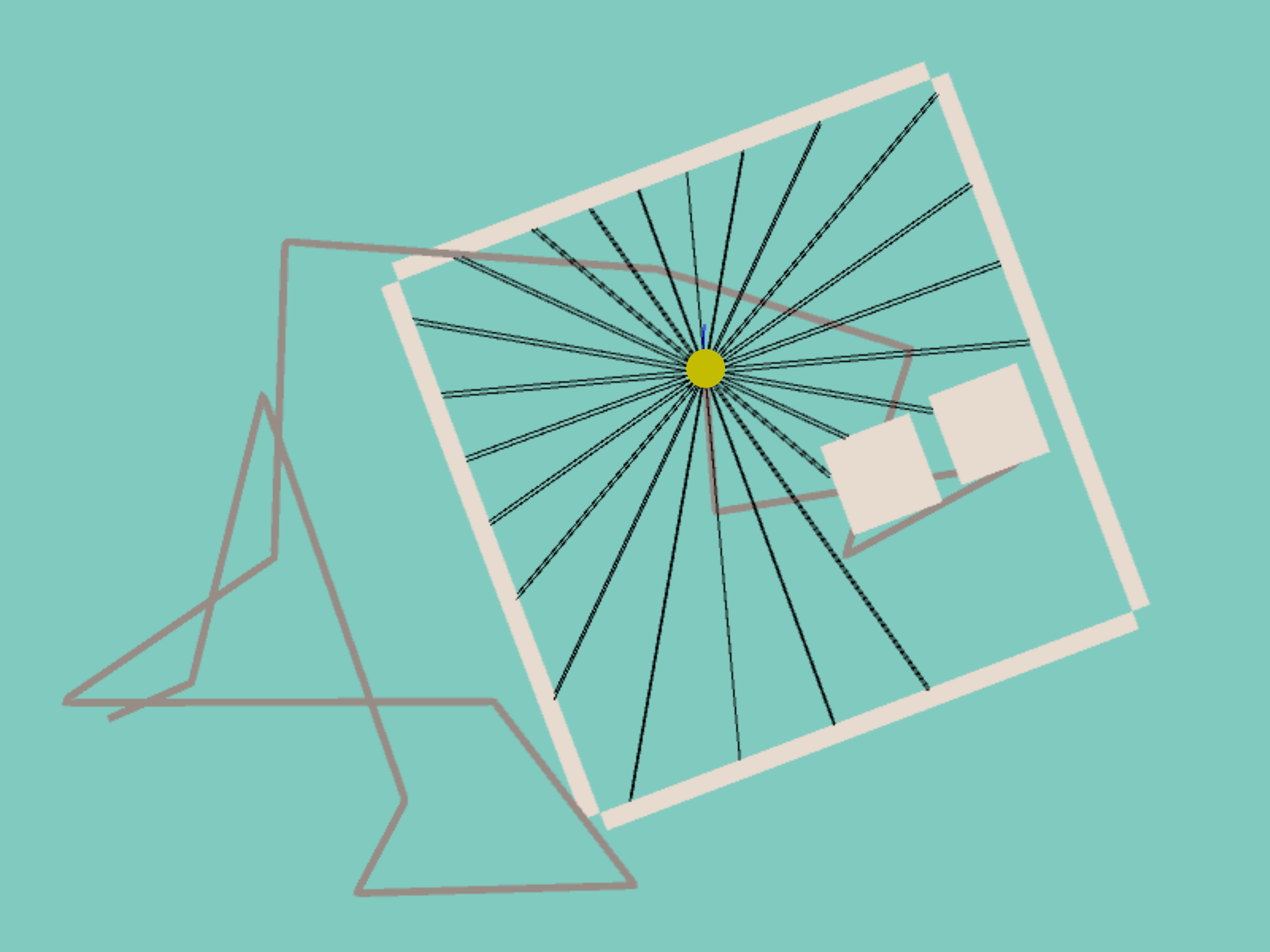}
 \caption{Training of the RL-based redirection controller. The yellow dots indicate the simulated user (agent). The cream frames and cubes represent the boundary of the tracking space and obstacles, respectively. The gray line represents the 100 m virtual walking path, and the black lines represent the rays measuring the surrounding distance. (In fact, although 60 rays were used, the number was reduced to 24 for better visibility.)}
 \label{fig:input}
\end{figure}

As inputs, we used position, orientation, distance to surrounding obstacles or boundaries in the \red{real space}, and the previous output vector.
The position in the tracked space was normalized between $-1$ and 1 according to half the length of the room, with the center of the room being 0.
Regarding the direction, the angle was normalized between $-1$ and 1 by dividing by $180^\circ$; here, the normal facing the center of the room was $0^\circ$.
The distances to the surrounding obstacles were observed every $6^\circ$ from the user's periphery, as shown in Fig.\ref{fig:input}, and the observed distances were normalized between 0 and 1 according to the length of the diagonal line of the tracking space.
The total input data corresponds to 64 vector observations, including two vectors for the $x$ and $z$ coordinates in the tracking space, one vector for the angle of the user with respect to the center of the tracking space, 60 vectors for the distances to obstacles or boundaries, and one vector for the previous output.

\subsection{Network Architecture and Training Settings}
\begin{figure}[tb]
 \centering 
 \includegraphics[width=0.8\columnwidth]{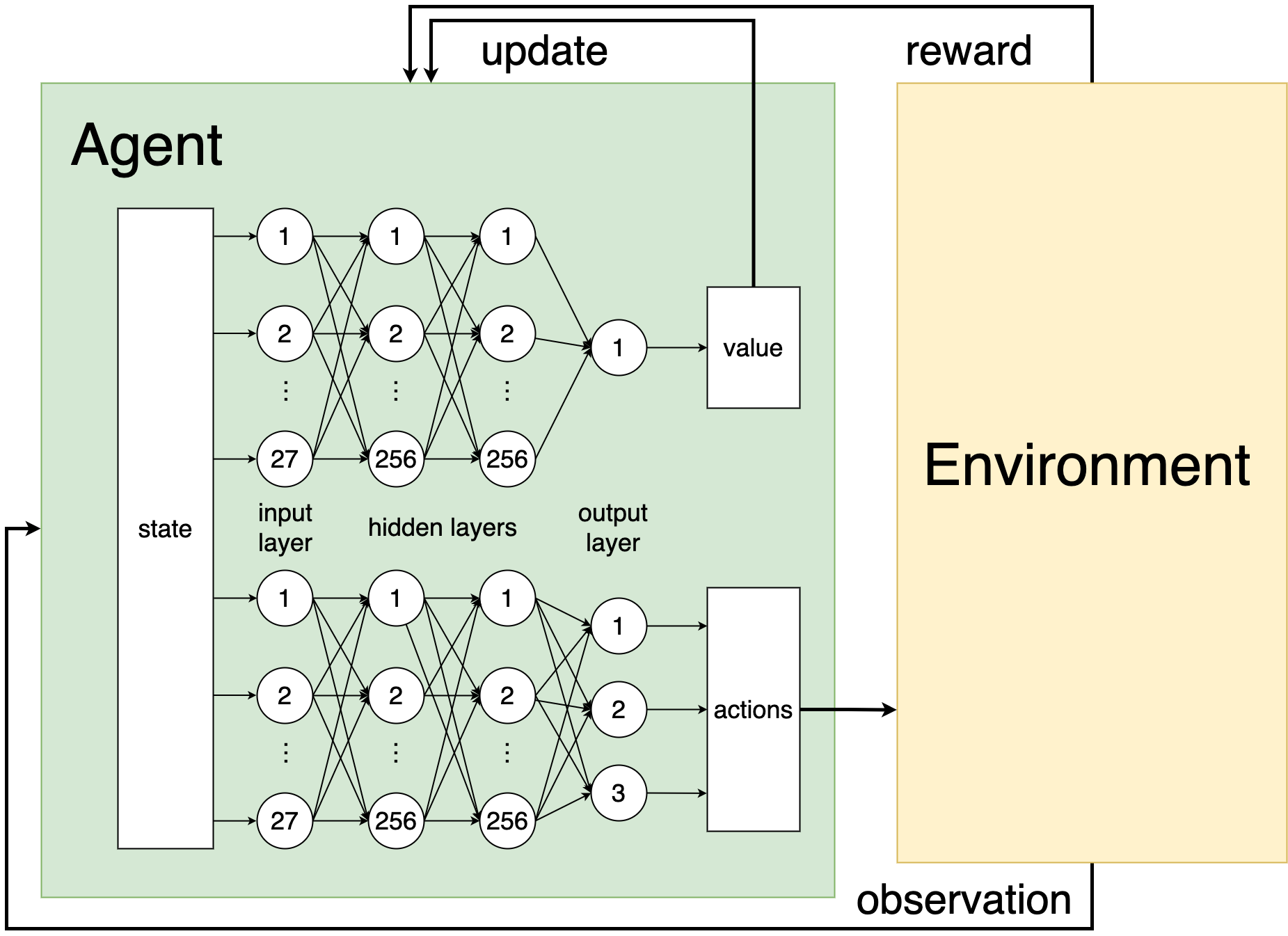}
 \caption{Model of proximal policy optimization used in our experiments. Sixteen agents were used during training and one during execution.}
 \label{fig:reinforcement}
\end{figure}

Fig.\ref{fig:reinforcement} shows the model of RL used in this experiment.
We used PPO \cite{Schulman2017} of the Unity ML-Agents Toolkit (Beta) \cite{juliani2018unity} \footnote{https://github.com/Unity-Technologies/ml-agents} for RL.
PPO maximizes a surrogate objective $\mathbb{E}[\rho_t(\theta)A_t]$, where $\rho_t(\theta) = \pi(a_t|s_t;\theta)/\pi(a_t|s_t;\theta_old)$ is the likelihood ratio of the recorded action between the updated and original policies. 
In PPO, the shared neural networks are updated in parallel by multiple agents.
The ML-agents toolkit was released by Unity Technologies \footnote{https://unity3d.com/} and is run using the Tensor-flow \footnote{https://www.tensorflow.org/} background.

We set rewards as shown in Table \ref{table:rewards}.
We normalized curvature gains ($g_C$) and translation gains ($g_T$) by using the maximal values of the gain, then squared it and multiplied it by $-0.01$.
The normalized gain was squared to suppress the sudden change of the gain.
We also added a negative reward of changing rate of curvature gains, noted as $\delta \ g_C$, to further suppress sudden changes.
For reset manipulation, we set $-45$ rewards for each resetting at the boundaries and obstacles.
In resetting, the rotation angle was not considered, and the reward was uniformly set.
To evaluate whether the current position of the agent is good or bad, we added an additional reward term, named {\it near obstacle punish}, that calculates the difference between the minimum and maximum distances towards the obstacles or boundary of the tracked space around the agent. Note that the {\it near obstacle punish} reward increases when the agent nears the center of the tracked space and decreases when the agent is near the boundary or obstacles.

\begin{table}[hbtp]
  \caption{Rewards of PPO}
  \label{table:rewards}
  \centering
  \begin{tabular}{lc}
  Parameter & Reward \\
  \hline
  Curvature gains &  $-0.01 * (|g_C - 1|/0.1333)^2$  \\
    &   $-0.1 * |\Delta g_C/0.1333|$ \\
  Translation gains & $-0.01 * (|g_T - 1|/0.26)^2$  \\
      &   $-0.1 * |\Delta g_T/0.26|$ \\
  Resetting  & $-45$ \\
  Near obstacle punish &  $0.2 * ( (d_{min}/d_{max}) -1)$ \\
  \hline
  \end{tabular}
\end{table}

The output normalized from $-1$ to 1 was multiplied by a coefficient and converted into translation gains, rotation angles, and curvature gains as follows.
To set the range of the translation gains within 0.86--1.26, the coefficient for the output of the translation gains was set to 0.2, and then 1.06 was added.
For the rotation angle, the output multiplied by 180 and modulo 360 was used.
To set the range of the curvature gains within $-$0.1333--0.1333, the coefficient for the curvature gains output was set to 0.1333.
Here, the positive and negative values indicate that the curvature is applied in the clockwise and counterclockwise directions, respectively.

The hyperparameter was tuned manually so that the reward would be maximized.
Please refer to the appendix for details on \red{the} hyperparameters.

With PPO, it is possible to learn in parallel using multiple agents. Thus, we used 16 agents for learning, and the total \red{number of} simulation steps was 16 million.

In RL, the learning time is generally inversely proportional to the CPU performance.
Using a PC equipped with an Intel Core i7-8750H CPU, the learning time of these models was approximately 2 h.
Additionally, the online execution time of the network was approximately 0.301 ms per frame, which is only 1.81\% of the duration of a frame in a 60 frames-per-second environment.
This execution time is sufficiently small to run the trained agent in practical VR contents with very low overhead.

\section{Environment Setups}

This section explicates common algorithms and the simulation frameworks of the experiments.
(The framework was used in experiments unless otherwise noted.)

\subsection{Tracked Space}

In this study, the tracked space was a square room with dimensions of 15 m $\times$ 15 m.
In experiments using obstacles, each obstacle was set as a cube with a side length of 2.5 m.
The obstacles were placed at random positions, and in case of multiple obstacles, it is possible for the obstacles to overlap.
In addition, the position of the obstacles was changed every 1000 frames, that is, every 100-m walk of the simulated user.

During the journey, the agent could move freely in the tracked space.
However, whenever the agent collided with walls or obstacles, the journey was interrupted and a reset procedure was performed, in order to reorient the agent. 

\subsection{Walking Agent}
To evaluate the performance of the proposed algorithms, a walking agent was designed to simulate how a human would walk in a virtual space.
At the start of the journey, a random target destination in the virtual environment was generated, and the agent selected the shortest path toward the target as its walking path.
In each simulation step, the agent will first moved 0.1 m along the walking path, and then new values of curvature and translation gains were applied according to the movement result.
After the agent has reached its target, the next target was generated immediately to create a continuous journey.
In all the following RL experiments of RL, we used 16 agents in parallel environments while training, and one agent while evaluating.

\subsection{Virtual Path Generation}
To allow the agents to move continuously, a new target would be generated in the VE would be generated whenever the agent reached the current target.
We used a procedural random path generator to make the agent move continuously during the journey.
When generating a new target, we considered the user's current position and proceeding direction as references, and set the target position according to different methods introduced in \cite{azmandian2015physical, azmandian2016redirected} (see Table \ref{table:path-generating-methods}), where $unif(a, b)$ is a random value from a uniform distribution between $a$ and $b$, and $Random$ is a combination of $Exploration\ small$ and $Exploration\ large$.

\begin{table}[hbtp]
  \caption{Path generating methods}
  \label{table:path-generating-methods}
  \centering
  \begin{tabular}{lcc}
    Category & Distance [m]  & Direction [rad]  \\
    \hline
    Office Building & $unif(2,8)$ & $\left\{ -\frac{\pi}{2}, \frac{\pi}{2} \right\}$ \\
    Exploration (small) & $unif(2, 6)$ & $unif(-\pi , \pi)$ \\
    Exploration (large) & $unif(8, 12)$ & $unif(-\pi , \pi)$ \\
    Long Walk & $\{1000\}$ & $unif(-\pi , \pi)$ \\
    Random & $unif(2, 12)$ & $unif(-\pi , \pi)$ \\
    \hline
  \end{tabular}
\end{table}

The maximal length of a journey was 100 km (one million simulation steps).
After reaching the maximal length, the agent terminated its journey, and no new target was generated.

\section{Preliminary Simulation: Heuristic Redirection Controller for Obstacle-based Environment}
Before testing the RL-based redirection controller, we investigated which combination of generalized controllers yields the best result in a multi-obstacle environment.

\subsection{Heuristic Algorithms}
\subsubsection{Translation Gains}

\begin{itemize}
\item Center-based Translation Gain
\end{itemize}
The {\it center-based translation gain (CTG)}, proposed by \cite{azmandian2015physical}, works as follows:

\begin{eqnarray}
g_{T} = \left\{
\begin{array}{ll}
1.26& ,\ if\ v_{center}\cdot v_{targ} < 0 \\
1& ,\ else
\end{array}
\right.
\end{eqnarray}

where $v_{center}$ and $v_{targ}$ are vectors pointing toward the center of the tracked space and in the direction in which the current user proceeds, respectively.
The main idea of CTG is to slow down the user when the user is moving away from the center of the tracked space.

\begin{itemize}
\item Advanced CTGs
\end{itemize}
Here, we propose the {\it Advanced CTG (ACTG)} algorithm, which is similar to CTG (See Algorithm \ref{alg1}).
When the user is moving away from the center of the tracked space, larger translation gains are applied to slow them down; similarly, smaller translation gains are applied when the user is heading toward the center of the tracked space. The unnoticeable minimum and maximum thresholds of translation gains were selected as 0.86 and 1.26, respectively, as introduced in \cite{steinicke2010estimation}.
Hence, the output translation gain of ACTG is calculated as follows:

\begin{equation}
g_{T} = 1.06 - 0.2 \cos{(\alpha_{center})}
\end{equation}

where $\alpha_{center}$ is the angle between the user's direction and the direction toward the center of the room, in \red{the range} $(-\pi, \pi )$.
This provides a smooth change in the translation gains when the user is changing direction.

\subsubsection{Reset Algorithms}

\begin{itemize}
\item Turn-to-Center
\end{itemize}
The 2:1-Turn is the most commonly used form of the reset. It performs a $180^{\circ}$ turn in reality, which in the VE is a $360^{\circ}$ turn \cite{Williams2007}.
However, as reported by Nguyen et al. \cite{Nguyen2018}, by using this method, the simulated user may go back and forth in the corners of the tracked space for a long time.
To overcome this problem, Peck et al. \cite{Peck2010} directed the user to the center of the tracked space at reset.
In addition, Nguyen et al. \cite{Nguyen2018} proposed a method to direct the user to the corner farthest from the user's position.
They named these methods as the to-center and to-corner methods, respectively \cite{Nguyen2018}.
As to-center may be mistaken for Steer-to-Center (S2C), we henceforth refer to to-center as {\it Turn-to-Center (T2C)}.

\begin{itemize}
\item Turn-to-Furthest
\end{itemize}
In addition to T2C and to-corner, we propose a new reset method called {\it Turn-to-Furthest (T2F)}.
Similar to to-corner, T2F forces the user to face in the direction of the furthest straight walking route from the user's current position.
As T2F considers the shape of the tracked space and the user's current position, it is expected to perform better than T2C and to-corner when there are random obstacles in the tracked space.

\subsubsection{Curvature Gains Algorithms}

\begin{itemize}
  \item S2C
\end{itemize}
{\it S2C} was chosen as the curvature gains algorithm because of its effectiveness in a generalized environment \cite{Hodgson2013}.
S2C simply steers the user to face the center of the tracked space by considering the center as most likely being the farthest place from walls and obstacles. 
We used the algorithm introduced in \cite{hodgson2011redirected}, which is a modification of Razzaque's original algorithm \cite{Razzaque2002} but more dynamic and versatile.

\begin{algorithm}                      
 \caption{ACTG}         
 \label{alg1}                          
 \begin{algorithmic}
 \renewcommand{\algorithmicrequire}{\textbf{Input:}}
 \renewcommand{\algorithmicensure}{\textbf{Output:}}
  \REQUIRE $P_{center/real}, P_{user/real}$
  \ENSURE $g_T$
  \STATE $\alpha \Leftarrow$ the angle between the user direction and the direction toward the center of the room, ranging in $(-\pi, \pi )$.
  \STATE $g_T \Leftarrow 1.06 -0.2 \cos{(\alpha)}$
  \RETURN $g_T$
 \end{algorithmic}
\end{algorithm}

\subsection{Simulation Setup}

We compared four different combinations of the heuristic controllers: the translation gain algorithm was chosen between CTG and ACTG, the reset algorithm was chosen between T2C and T2F, and the S2C was selected as the curvature gains algorithm.
We simulated each controller in a 100-km journey (one million simulation steps) with 0--3 obstacles.
Random path generation was used for the walking path in the virtual space, and we used the count of the reset procedure as the evaluation criterion.

\subsection{Results}
\begin{figure}[hbtp]
 \centering 
 \includegraphics[width=\columnwidth]{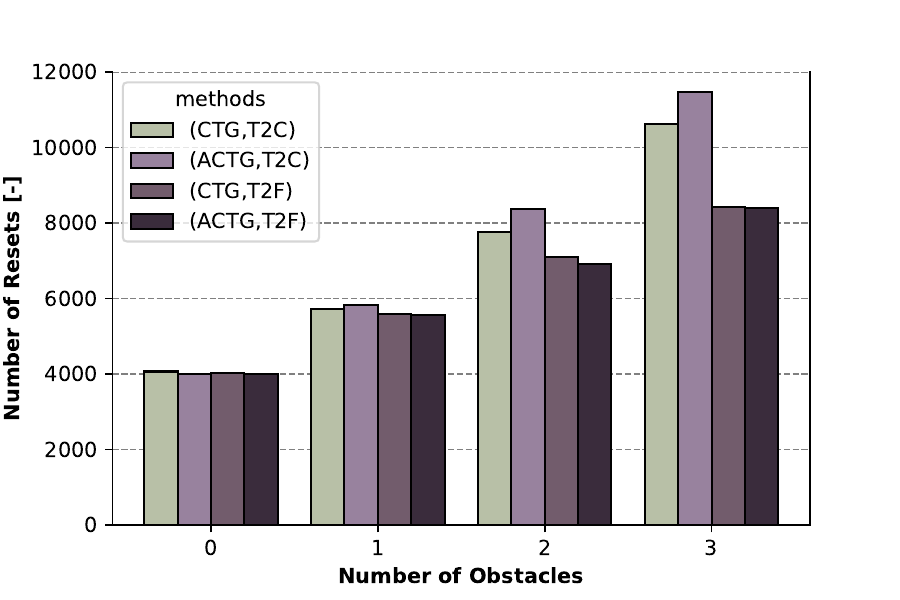}
 \caption{Number of resets under the multiple obstacles environment}
 \label{fig:resetCountGraph_exp0}
\end{figure}

Fig. \ref{fig:resetCountGraph_exp0} shows the number of resets in each heuristic condition with obstacles.
As shown, the number of resets increases with the number of obstacles in all methods, as the reset count is doubled when three obstacles are placed.
No large difference was observed in the zero-obstacle condition; however, in the three-obstacle condition, the methods using T2F performed $20.5\%$ (CTG) and $26.8\%$ (ACTG) better than those using T2C.
Among the four conditions, the method with the \red{fewest} resets in any \red{real environment} was that using ACTG, T2F, and S2C.
Henceforth, the heuristic condition is referred to as (ACTG, T2F, S2C).

\subsection{Discussion}

We found that adopting T2F as a reset algorithm in obstacle-based environments reduces the number of resets more than when using T2C.
This could be because T2C cannot avoid obstacles near the center of the \red{real space}, whereas T2F finds the furthest point to turn, other than the center.
In addition, the comparison of (CTG, T2F) and (ACTG, T2F) indicated that the combination of (ACTG, T2F) \red{resulted in fewer} resets in all environments.
Therefore, the most effective conventional generalized heuristic controller is the one using the combination of (ACTG, T2F, S2C).
Hereafter, this best-performing heuristic controller is called a {\it heuristic controller} and is compared with other controllers using RL.

\section{Simulation 1: Reinforcement Learning for Translation, Rotation, Curvature Gains}

To evaluate the effectiveness of RL, we replaced the translation gains, reset angles, and curvature gains algorithm in the heuristic algorithm combination with RL individually.
We trained and tested the training result against the heuristic algorithm combination.

\subsection{Simulation Setup}
We \red{simulated} each algorithm combination in a 100-km journey (one million simulation steps) for training and evaluation by using random path generation, and recorded the total count of resets.
No obstacle was placed in the tracked space.

Table \ref{table:usedMethod_exp1} shows the algorithm used for each redirection controller. The Heuristic controller used (S2C, T2F, ACTG), which was the best result in the preliminary experiment. RL-prefixed controllers replaced one of the curvature, translation, and reset algorithms from the heuristic with RL, respectively.

\begin{table}[hbtp]
  \caption{Algorithm used}
  \label{table:usedMethod_exp1}
  \centering
  \begin{tabular}{lccc}
    & Translation  & Reset  & Curvature  \\
    \hline
    Heuristic & ACTG & T2F & S2C \\
    RL-translation  & RL & T2F & S2C \\
    RL-reset  & ACTG & RL & S2C \\
    RL-curvature  & ACTG & T2F & RL \\
    \hline
  \end{tabular}
\end{table}

\subsection{Results}

Table \ref{table:result_exp1} shows the number of resets in the heuristic condition, RL-curvature condition, RL-reset condition, and RL-translation condition.
The RL-reset and RL-translation conditions were slightly less effective than the heuristic condition, although the RL-curvature condition had almost the same number of resets as that in the heuristic condition.

\begin{table}[hbtp]
  \caption{Number of Resets in Experiment 1}
  \label{table:result_exp1}
  \centering
  \begin{tabular}{lcccc}
    Controller & Number of Reset  \\
    \hline
    Heuristic & 3994  \\
    RL-curvature & 3996  \\
    RL-reset & 4226  \\
    RL-translation & 4152  \\
    \hline
  \end{tabular}
\end{table}

\begin{figure}[tb]
 \centering
 \begin{tabular}{c}
 \begin{minipage}[c]{\linewidth}
  \centering
  \includegraphics[width=0.9\columnwidth]{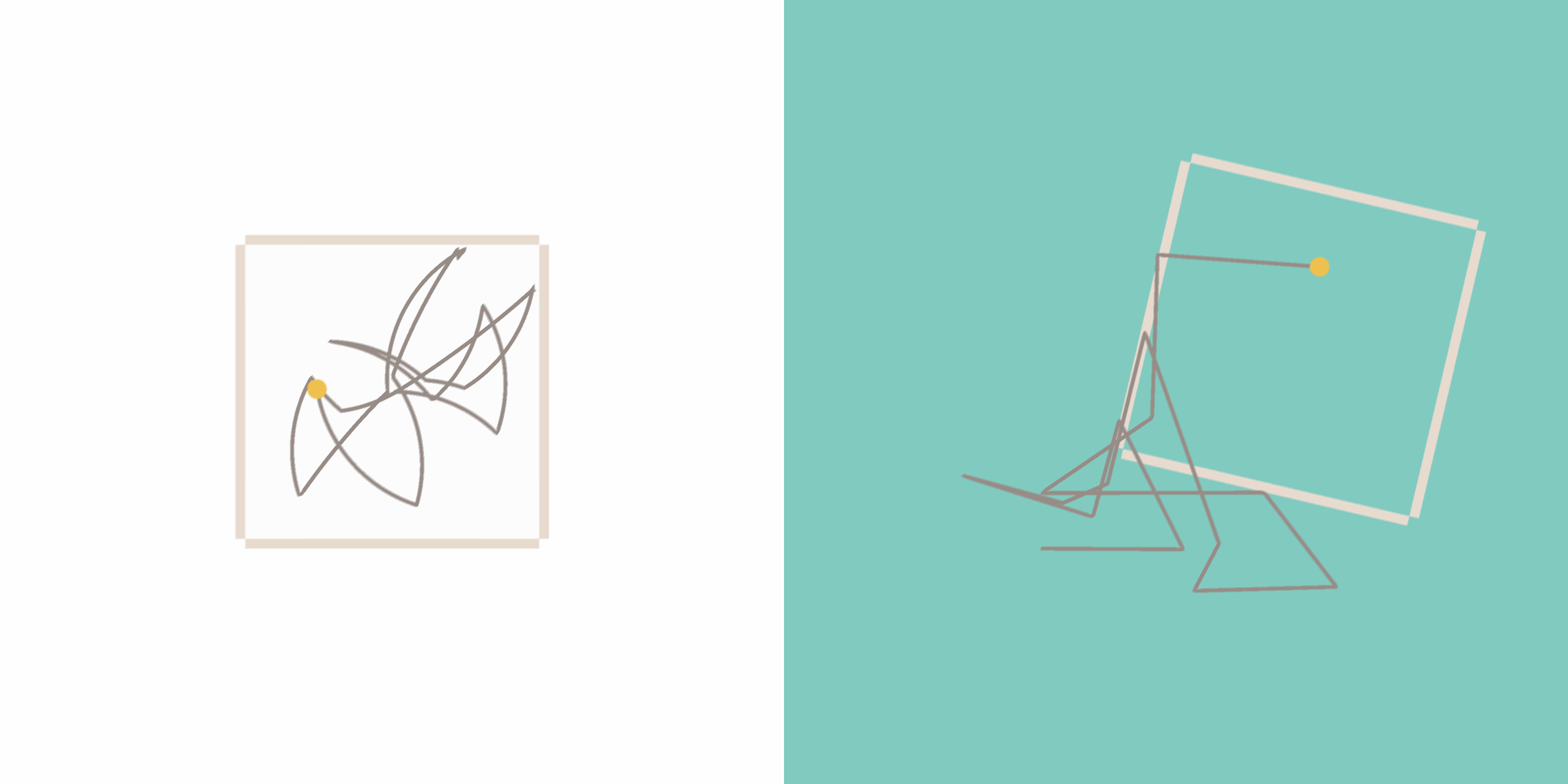}
  \subcaption{Heuristic controller}
 \end{minipage}\\
 \begin{minipage}[c]{\linewidth}
  \centering
  \includegraphics[width=0.9\columnwidth]{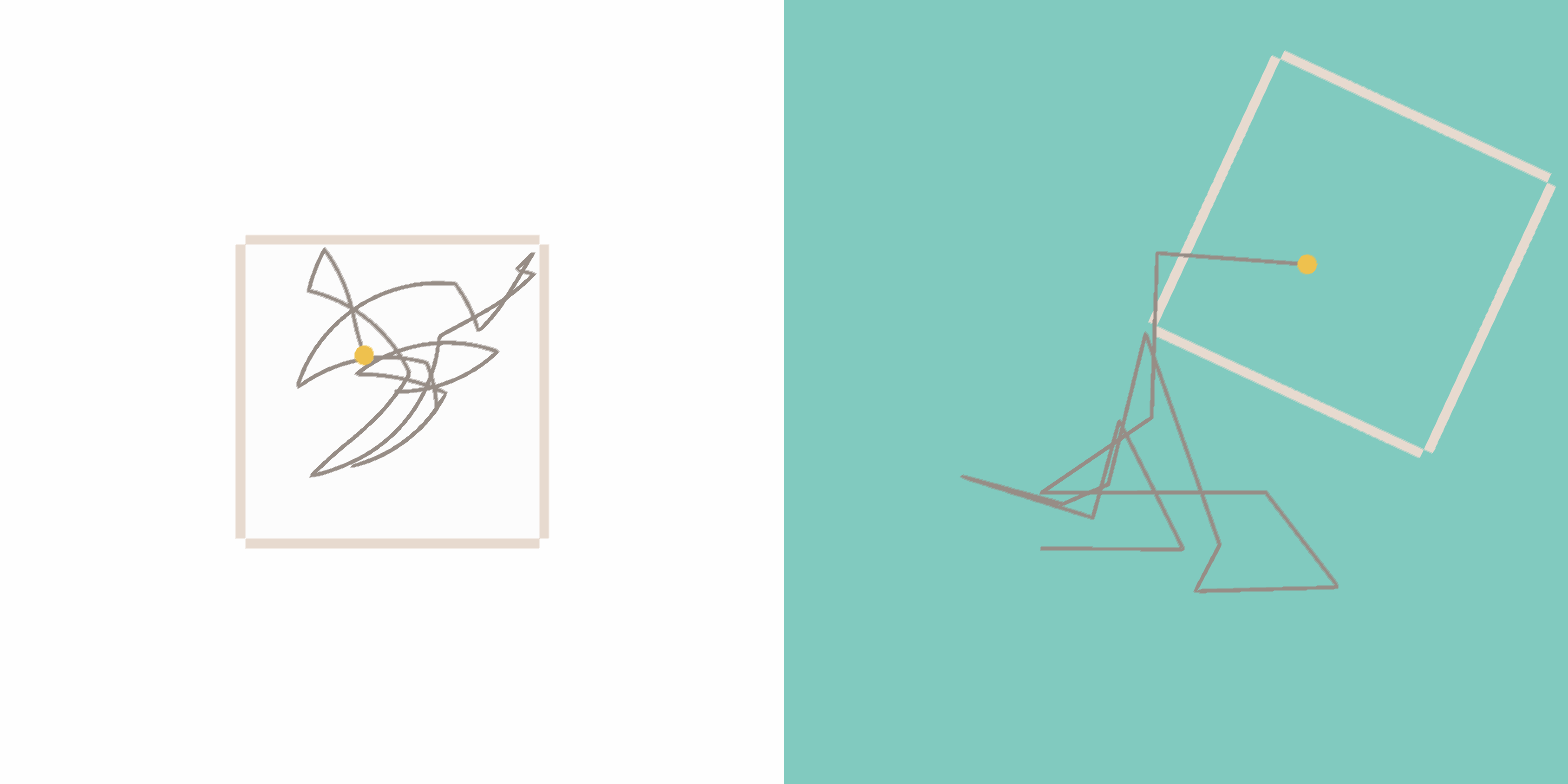}
  \subcaption{RL-curvature controller}
 \end{minipage}
 \end{tabular}
 \caption{Walking path of the initial 100 m when using the heuristic and RL-curvature controllers. The left-hand side shows the \red{real space} and the right-hand side shows the virtual space. The yellow dots indicate simulated users (agents). The cream frames represent the boundary of the tracking space, and the gray lines represent the 100 m physical and virtual walking paths, respectively.}
 \label{fig:walkingPath}
\end{figure}

Fig. \ref{fig:walkingPath} shows the walking paths of the initial 100 m when using the heuristic and RL-curvature controllers.
As shown, the RL-curvature controller \red{exhibited} a behavior similar to those of S2C and Steer-to-Orbit.

\begin{figure}[tb]
 \centering 
 \includegraphics[width=\columnwidth]{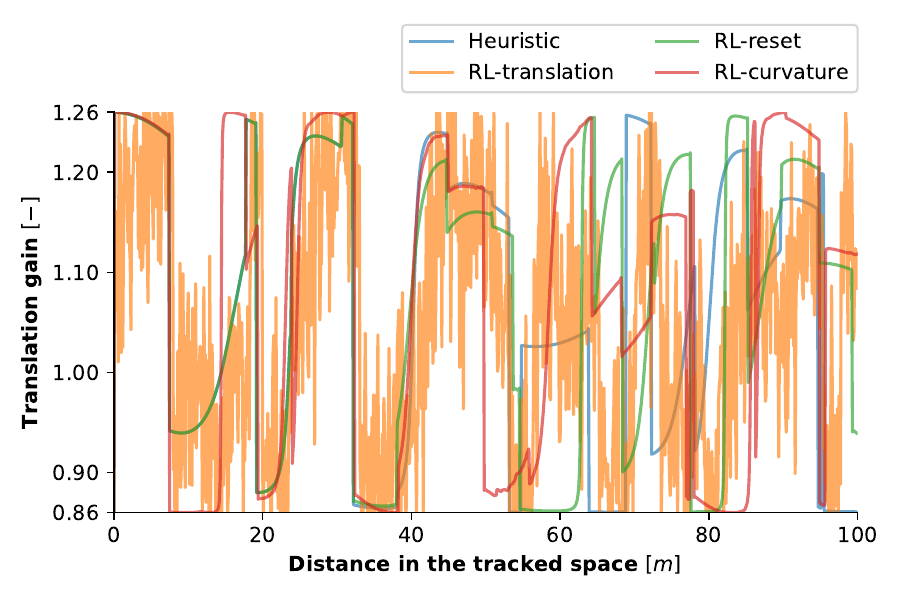}
 \caption{Translation gains when a simulated user is walking the initial 100 m in the \red{real space}.}
 \label{fig:exp1_trans}
\end{figure}

Fig. \ref{fig:exp1_trans} shows the translation gains when a simulated user is walking the initial 100 m in the \red{real space}.
In Heuristic, partial RL-reset, and partial RL-curvature algorithms, translation gains changed discontinuously, whereas in the partial RL-translation algorithm, translation gains oscillated but changed continuously.
The same pattern was observed in the initial 100 m and beyond.

\begin{figure}[tb]
 \centering 
 \includegraphics[width=\columnwidth]{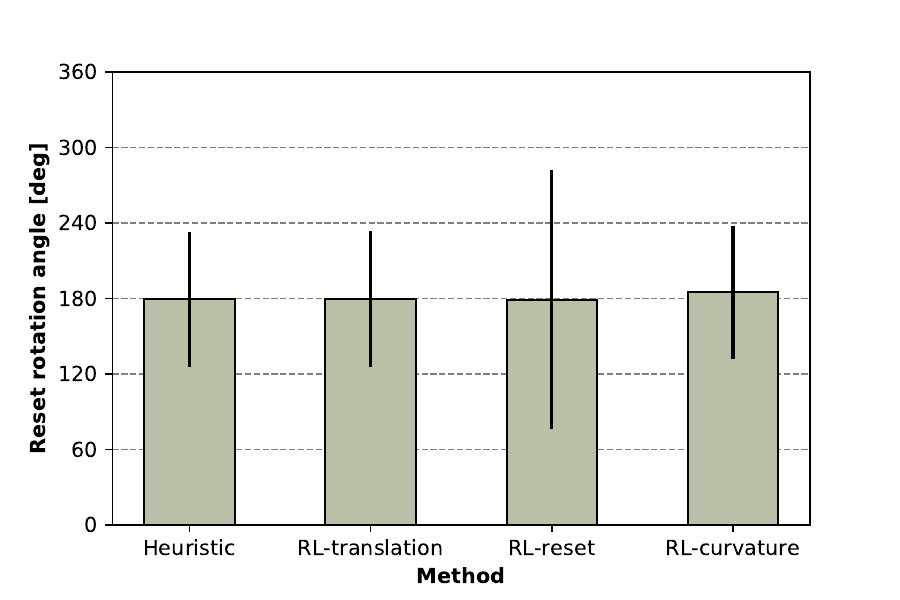}
 \caption{Average and standard deviation of reset angles $(mean \pm SD)$}
 \label{fig:exp1_reset_bar}
\end{figure}

Fig. \ref{fig:exp1_reset_bar} shows the average reset angles when a simulated user is walking 100 km in the \red{real space}. The error bars indicate standard variance.
In all algorithms, the average reset angles were approximately $180^{\circ}$, whereas  the variance of the reset angle of the RL-reset algorithm was greater than that of the other algorithms.

\begin{figure}[tb]
 \centering 
 \includegraphics[width=\columnwidth]{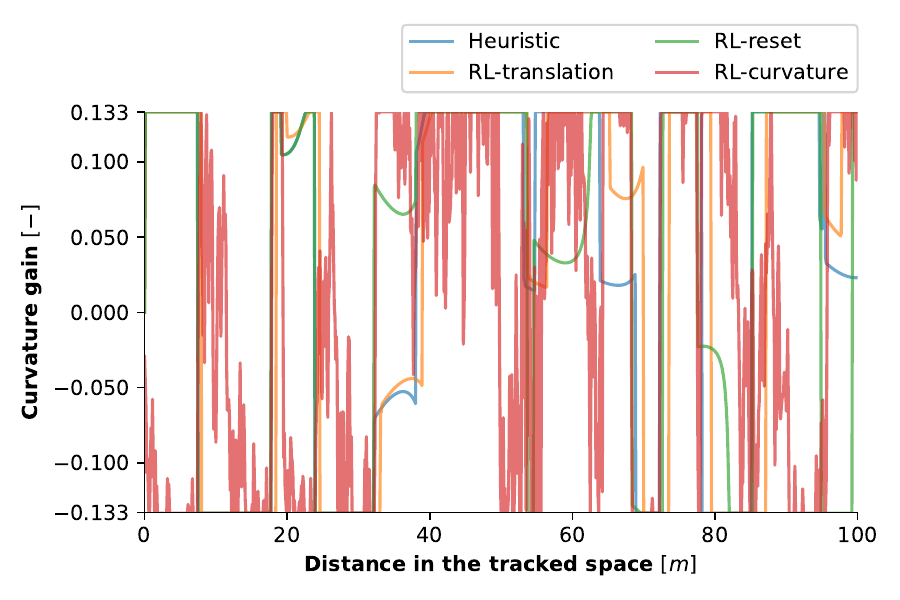}
 \caption{Curvature gains when a simulated user is walking the initial 100 m in the \red{real space}.}
 \label{fig:exp1_curv}
\end{figure}

Fig. \ref{fig:exp1_curv} shows the curvature gains when a simulated user is walking the initial 100 m in the \red{real space}.
In the Heuristic, partial RL-translation, and partial RL-reset algorithms, the curvature gains change discontinuously, whereas in the partial RL-curvature algorithm, the curvature gains oscillated but changed continuously.
The same pattern was observed in the initial 100 m and beyond.

\subsection{Discussion}

In the unobstructed environment, there were fewer resets for the heuristic controller than for the RL-translation and RL-reset controllers.
In contrast, the number of resets was almost the same in the heuristic condition in which RL was applied to the curvature gain.

When we applied RL to translation gains and curvature gains, the output gains oscillated.
The application of such a sudden change in gains to an actual user may cause awareness of operations and VR sickness to the user.
In contrast, when the RL-curvature controller was applied, the walking path in the \red{real space} was smooth and did not show any irregularity.
The effects of these gain oscillations on the user are discussed in Section IX: User Study.

Hereinafter, among the controllers using RL, the RL-curvature controller with the smallest number of resets is called an {\it RL controller}.

\section{Simulation 2: RL-based Redirection Controller for Obstacle-based Environments}
Often there are various obstacles in a \red{real space}.
For example, furniture---such as tables and sofas---and pillars can be considered as obstacles.
These obstacles can have a significant impact on the redirection controller, as discussed in Section V.
In this section, we compare the heuristic redirection controller and the redirection controller using RL in the \red{real space} with obstacles.

\subsection{Simulation Setup}
To verify the versatility of the RL-trained redirection controllers in obstacle-based environments, we placed obstacles in the tracked space and evaluated their results.
We compared the three conditions of the heuristic controller (heuristic), the RL redirection controller trained in Experiment 1 (non-retrained), and the RL redirection controller retrained in the environment with obstacles (retrained).
In the non-retrained model, the controller trained in the \red{real environment} without obstacles was used to evaluate the behavior in the \red{real environment} with $n$ obstacles.
In the retrained model, the controller trained in the \red{real environment} with $n$ obstacles was used to evaluate the behavior in the \red{real environment} with $n$ obstacles.

We ran each algorithm in a 100-km journey (one million simulation steps) for training and evaluation by using the random path generation, and recorded the total count of resets in the evaluation.

\subsection{Results}

\begin{figure}[tbp]
 \centering
 \includegraphics[width=\columnwidth]{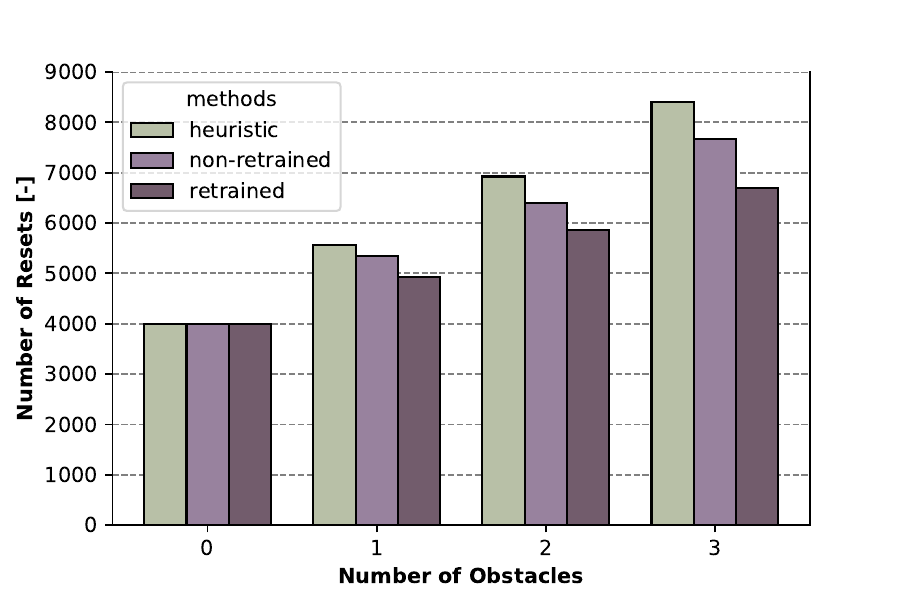}
 \caption{Number of resets in Simulation 2}
 \label{fig:resetCountGraph_exp2}
\end{figure}

Fig. \ref{fig:resetCountGraph_exp2} shows the total count of resets in the heuristic, non-retrained RL, and retrained RL conditions. One can observe that although the performance of the three controllers was almost the same when there were no obstacles, it differed considerably when obstacles were added.
When using the heuristic controller in environments with one, two and three obstacles, the number of resets increased by 39.4\%, 73.4\%, and 110\%, respectively, in comparison to an environment without obstacles.
When using the RL controller trained in Experiment 1, the relative increment in the number of resets was 33.9\%, 60.3\%, and 91.7\%, respectively, and the increment in the number of resets when using the retrained RL controller was 23.3\%, 46.6\%, and 67.4\%, respectively.
As observed, in environments with the same number of obstacles, both RL controllers had a lower increment of resets than the heuristic controller.
Therefore, in the presence of obstacles, the RL controller outperformed the heuristic controller with both the non-retrained model in Experiment 1 and the retrained model in each environment with each path-generating method.

\subsection{Discussion}

Environments with obstacles are more similar to a real environment than a plain environment.
Therefore, it can be said that the method that performs better in environments with obstacles is more versatile.
Our results show that the RL-based redirection controllers are advantageous in \red{real spaces} with obstacles, as the number of resets is increases at a much slower pace than that when using heuristic redirection controllers for an increasing number of obstacles.
Even when the controller is not trained in an environment with obstacles, it somehow performs better than the heuristic controller, showing that the RL-based redirection controller has general versatility over environments \red{with different shapes}.

\section{Simulation 3: Learning In Different Walking Paths}
In Simulations 1 and 2, we investigated the effect of the spatial configuration of the real space on the redirection controller using RL.
In this section, we investigate how walking paths in a VE affect the result of the RL-based redirection controller.
We used five virtual paths and three controllers to evaluate the effectiveness of the RL-based redirection controller.

\subsection{Simulation Setup}
We compared path-generating methods of $Exploration\ Small$, $Office$, $Exploration\ Large$, and $Long\ Walk$ (Table \ref{table:path-generating-methods}) described in Section IV.C by using the heuristic redirection controller, the RL redirection controller trained in Experiment 1, and the RL redirection controller retrained in the environment with the various walking paths.
In this simulation, no obstacle was placed in the tracked space.
We \red{simulated} these controllers in a 100-km journey (one million simulation steps) for evaluation, and recorded the total count of the reset procedures.

\subsection{Results}

\begin{figure}[tbp]
 \centering 
 \includegraphics[width=\columnwidth]{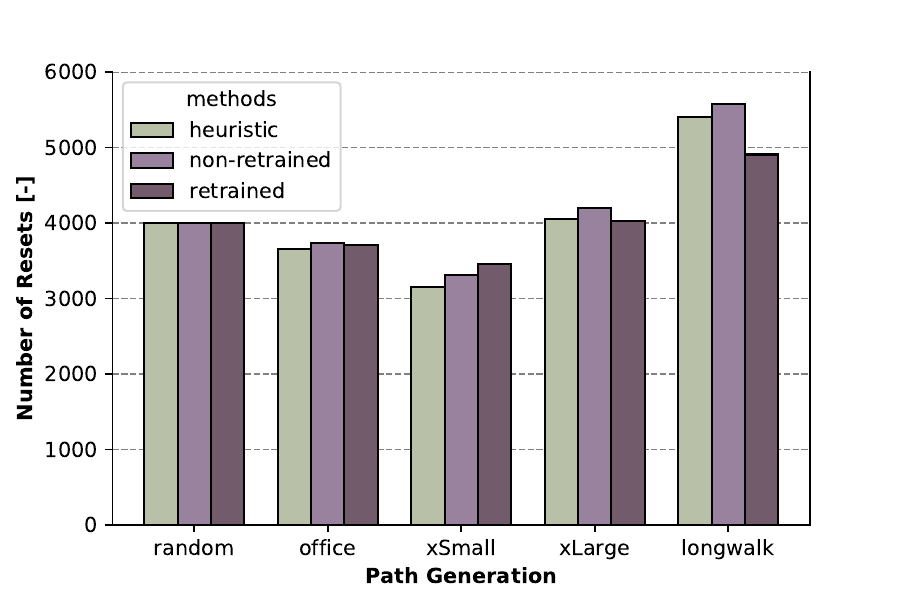}
 \caption{Number of resets in Simulation 3}
 \label{fig:resetCountGraph_exp3}
\end{figure}

Fig. \ref{fig:resetCountGraph_exp3} shows the total count of resets in the four different path-generating algorithm conditions.
As shown, the number of resets differs considerably in each path-generating algorithm.
In comparison to heuristic, the RL method performs better in the Long Walk condition, slightly better in the Exploration (large) condition, and worse in the Office and Exploration (small) conditions.

\subsection{Discussion}

As shown in Table \ref{table:path-generating-methods} and Fig. \ref{fig:resetCountGraph_exp3}, the number of resets increases when the average distance between targets increases.
Further, we found that the effectiveness of the RL-based redirection controller differs depending on the path-generating methods.
Evaluated in comparison to the heuristic redirection controller, the effectiveness in ascending order is as follows: Exploration (small), Office, Random, Exploration (large), Long Walk, which is also the average distance between targets in ascending order.

We assumed that the difference in the effectiveness is due to the average distance between targets, which may affect the efficiency of learning. As evidence, the newly trained model in Exploration (small) actually performed worse than the non-retrained model, which is an unexpected phenomenon.
This requires further verification of the relation between average distance and learning efficiency.

\section{User Study}

In this experiment, a demonstration produced using the learned RL curvature gain controller of Simulation 2 and the heuristic controller was used to compare the performances of the RL controller and the heuristic controller under different real-world conditions.
To verify the effectiveness of the RL and heuristic controllers in simple and complex real environments, the experiment was conducted in a within-subject design for two controller conditions (the heuristic controller and the RL controller) $\times$ two real environmental conditions (no-obstacle environment and with-obstacle environment), and the number of resets was recorded.
In this user study, we also investigated the effect of gain oscillation on the user's VR sickness, which is caused by the oscillation of the curvature gain when using the RL controller as confirmed by the simulation.
The duration of the experiment per person was approximately 1 h, including the briefing, pre-questionnaire, trials, and post-questionnaire.
This user study was approved by the local ethics committee.

\subsection{Participants}
Participants were recruited from within and outside the local campus through Twitter.
A total of 24 participants (11 females, 13 males; mean age: 25.4 years, SD: 5.38) were recruited for the user study.
Among them, 10 users played 3D games more than once a month, and 8 users experienced VR more than once a month.
None of the participants reported any visual impairment.
The participants were paid 1,000 JPY (approximately 9 USD) via Amazon Gift Card.

\subsection{\red{Experimental Environment}}
We performed all trials in an 8 m $\times$ 8 m tracked space.
\red{For the with-obstacle environment condition, we did not place any actual obstacles in the real space to prevent users from being injured by collisions, but we set up two off-limit areas of 1.5 $times$ 1.5 m each, corresponding to the obstacles.}

During the user study, participants wore an HTC Vive Pro HMD connected to a portable backpack PC (MSI VR One with Intel Core i7 and GeForce GTX 1060), and followed instructions to walk through a specified virtual path.

\begin{figure}[tbp]
 \centering
 \includegraphics[width=\columnwidth]{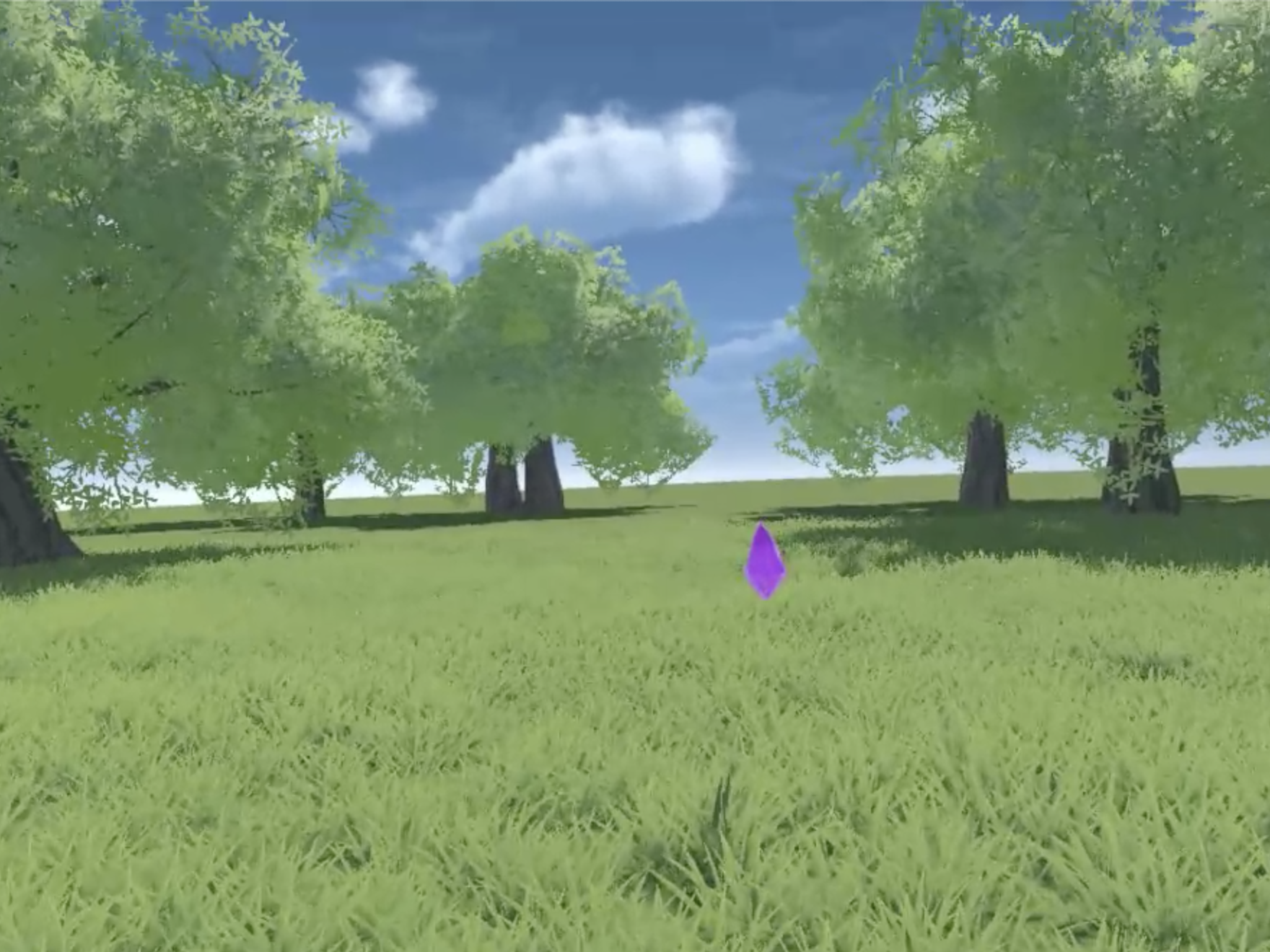}
 \caption{Virtual environment used in the user study. Participants were asked to collect purple crystals.}
 \label{fig:fpv}
\end{figure}

The VR environment was a grassy field with a small mountain and \red{some trees were placed in the background to make the user aware of the relative position} (see Fig. \ref{fig:fpv}.)
In the VR environment, crystals were displayed one at a time at a height of 1.2--1.5 m above the ground, and users were instructed to walk toward the crystals to retrieve them.
The position of each crystal was set so that the user's walking path was a {\it Random} path in the simulation conditions.

In the simulation environment, the reset was instantaneous; however, in the user study, the reset was performed using a modified 2:1-turn method, which is based on the 2:1-turn method \cite{10.1145/1272582.1272590}.
A reset was triggered when the user entered an area within 0.5 m of the obstacle or the boundary of the tracked space, and the user was supported to rotate $360^\circ$ on the spot in the VR space.
During this time, the user's direction of travel was controlled by applying a rotation gain that rotated $360^\circ$ to $540^\circ$ in real space.

\subsection{\red{Experimental Procedure}}
The participants received an overview of the user study and completed a consent form.
They were then asked to fill out a preliminary questionnaire regarding their sex, age, height, visual impairment, frequency of playing 3D games, and frequency of using VR.
On completing the questionnaire, the participants were briefed on how to use the equipment, and then they were fitted with the equipment.


First, the users performed a practice session on a 25 m walking path.
No redirection was performed during the practice session and only resetting was applied.
Then, four conditions of the 100 m walking path in reality were performed.
We counterbalanced order effects by presenting the four conditions in a different order for each participant.

Before and after each trial, the participants completed the Simulator Sickness Questionnaire (SSQ) \cite{doi:10.1207/s15327108ijap0303}.
After completing all the trials, the participants completed the iGroup Presence Questionnaire (IPQ) \cite{6790846}.
In addition, an open-ended questionnaire on the experiment was conducted orally after the experiment.

\subsection{Results}

\begin{figure}[tbp]
 \centering
 \includegraphics[width=\columnwidth]{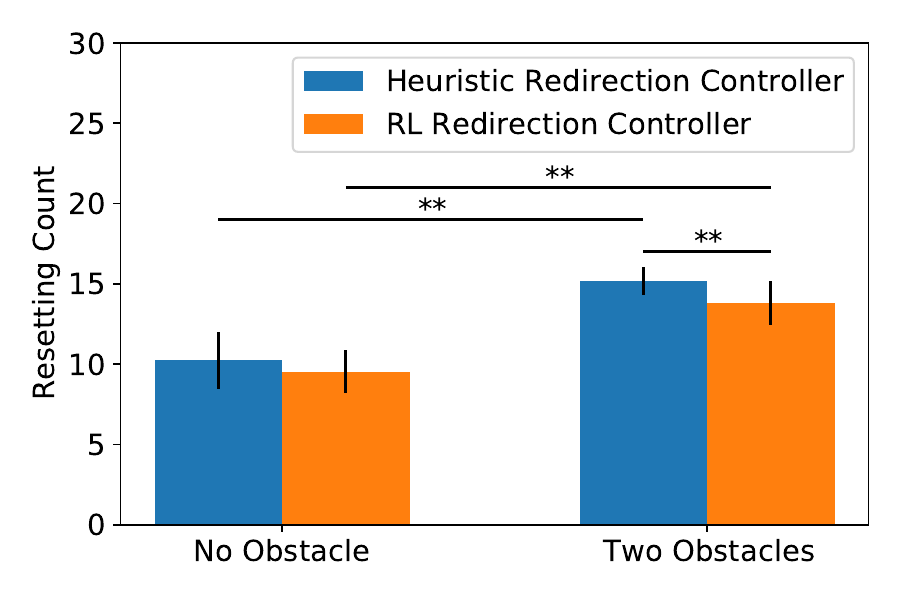}
 \caption{Resetting counts in the user study. The error bars show standard deviation. **: p < 0.01.}
 \label{fig:userstudy}
\end{figure}

\begin{figure}[tbp]
 \centering
 \includegraphics[width=\columnwidth]{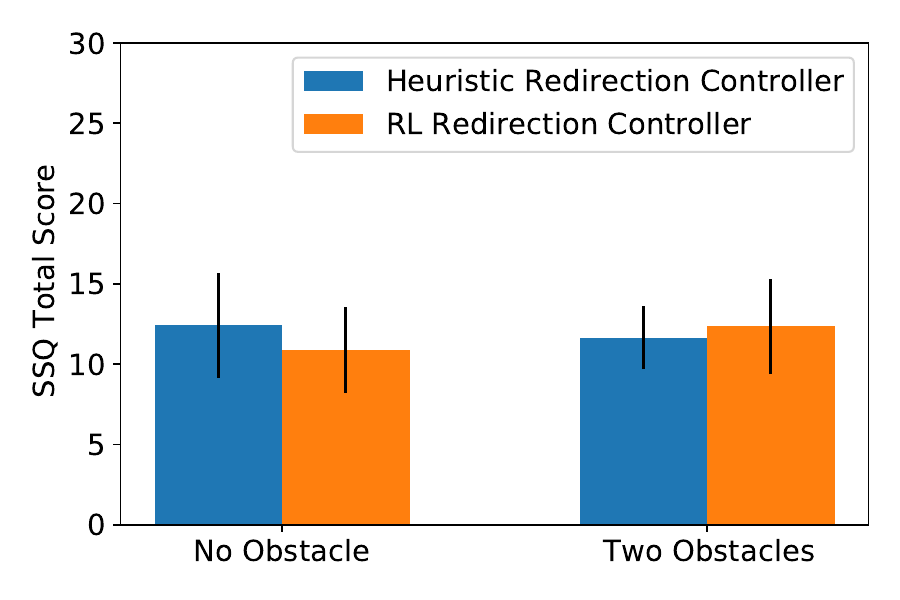}
 \caption{SSQ total score in the user study. The error bars show standard error.}
 \label{fig:ssq}
\end{figure}

Fig. \ref{fig:userstudy} shows the mean and standard deviation of the reset counts for each condition.
The results of two-way analysis of variance (ANOVA) were as follows:
Environments: F(1,23) = 187.83, p < 0.01; Controller: F(1,23) = 57.14, p < 0.01; and Environments $\times$ Controller interaction: F(1,23) = 22.69, p < 0.01.
The post hoc Holm-corrected t tests found that the number of resets under the two obstacle environments differed between the heuristic and RL controllers: F(1,23) = 111.75, p < 0.01, and that each controller had a different number of resets under different environments (heuristic controller: F(1,23) = 139.24, p < 0.01, RL controller: F(1,23) = 32.45, p < 0.01).

Fig. \ref{fig:ssq} shows the mean and standard error of the SSQ total score for each condition.
From the results of two-way ANOVA, no significant effect was observed.

The results of the IPQ showed that the overall rating of presence for all participants was 3.38/6 (SD = 0.767).
The results for the three subscales of the IPQ were as follows: Spatial Presence (M = 3.94/6, SD = 0.874) and Involvement (M = 3.22/6, SD = 1.45) were moderately high, while Realness (M = 2.53/6, SD = 0.922) ) was low.

\subsection{Discussion}
The results of the user study showed that the number of resets was significantly increased regardless of the controller type in a complex environment with obstacles.
\red{However}, the RL controller tended to reset less often than the heuristic controller in all environments, especially in the two-obstacles environment.
These results are similar to those obtained in Simulation 2, indicating the superiority of the RL redirection controller in complex environments.

\section{General Discussion}

Through Simulations 1--3 and the user study, we determined that our proposed RL controller has general versatility and can perform better in environments with complex tracked spaces, compared with the heuristic controller.

In Simulation 1, the translation gain, reset angle, and curvature gain algorithms in the heuristic algorithm combination were individually replaced by RL to verify the manipulations for which RL is suitable.
Among the three redirection controllers with a single component replaced with RL, the RL-curvature controller that replaced the curvature gain with RL control showed the best performance and was almost as effective as the heuristic controller.
However, problems such as oscillation of the gain were observed in the controller using RL, which may lead to discomfort and VR sickness. This issue was verified in the user study described below.

In Simulation 2, we compared the heuristic redirection controller and the redirection controller using RL in the \red{real space} with obstacles.
When comparing the heuristic method, the model trained in an obstacle-free environment, and the model trained in an obstacle environment, the model trained in an obstacle environment showed the best performance.
This trend increased as the number of obstacles in the real space increased.
Furthermore, the model trained in the obstacle-free environment also performed better than the heuristic method in the obstacle environment.
Therefore, it is suggested that the proposed RL controller is versatile in that the model learned in a specific environment can be used in different environments.

In Simulation 3, we investigated how different walking paths affect the result of the RL-based redirection controller.
As the distance between the targets increased, the effectiveness of the retrained RL controller tended to outperform the heuristic controller.
Interestingly, in some situations, the retrained models showed worse results than the non-retrained models and heuristics.
This shows that the proposed model is suitable for application to some walking paths.

In the user study, we examined the effectiveness of the proposed RL controller using the demonstration created through the proposed controllers.
The results of the user study show that the proposed RL controller significantly outperforms the heuristic controller in the presence of obstacles in real space.
In addition, as mentioned earlier, we applied the output to the gain values; however, it can be observed that the output value is highly unstable.
Zhang et al. \cite{Zhang2014, Zhang} investigated the effect of dynamically changing the translation and rotation gains and concluded that the oscillations of these gains lead to VR sickness and awareness of redirection operation.
Therefore, the oscillation of the curvature gain is thought to lead to VR sickness and awareness of redirection manipulation.
From the experimental results, there was no difference in the SSQ score, which is an index of VR sickness, between the proposed method and the existing method.
This suggests that the oscillation of the gain associated with the use of the RL controller did not lead to VR sickness.

These results indicate that the proposed RL controller outperforms the conventional heuristic controller in a real space with many obstacles.
They also suggest that the model trained in only one real environment outperforms the heuristic controller, although the performance is not as good as the controller retrained in a different real environment. 
In contrast, depending on the walking path in the VR environment, the performance of the RL controller may be lower than that of the conventional heuristic controller.
This suggests that the proposed RL controller has a certain degree of versatility for real environments, but is path-specific for VEs.

In this study, we analyzed methods that replace translation, rotation, and curvature gains with RL.
Various combinations of these methods may improve the performance of the controller.
Moreover, in this study, we assumed that the obstacle in the real environment was stationary, but we aim to study dynamic obstacles in the future.
If the effectiveness of the proposed controller could be confirmed for dynamic obstacles, it would be possible to apply the controller to multiple users in the same real environment by considering the other users as dynamic obstacles.
Additionally, in this study, we examined the effects of the spatial configuration of the real space and the walking path in the VR space on the effectiveness of the proposed RL RDW controller; we would like to examine the effects of the contents in the VR space on the RL RDW controller in the future.
Furthermore, we used the number of resets as an evaluation metric in the simulation, but another possible evaluation metric could be the ease of noticing the RDW manipulation and the occurrence of VR sickness.
These indices have been verified through user studies; however, it is difficult to verify the performance of a controller with many parameters, such as a controller using machine learning, through user studies.
To verify the ease of noticing the operation in the simulation, a virtual user that takes into account the human perception model and behavior model might be considered in the future.

\section{Conclusion}

This paper presented a novel redirection controller using RL that can operate online.
The proposed controller can create an optimal plan according to the spatial composition of real spaces and VEs.
Moreover, it can cope with obstacles placed in the real space.

We quantitatively evaluated heuristic and proposed redirection controllers through simulations and a user study.
In Simulation 1, the effectiveness of the redirection controller using RL was verified by applying RL to a part of the heuristic redirection controller.
The results of the simulation experiment demonstrated that the redirection controller with curvature gain controlled using RL was able to reduce the number of resets by 20.3\% against the best heuristic controller.
In addition, despite no prior knowledge, the RL controller \red{exhibited} a behavior similar to those of the S2C and Steer-to-Orbit. However, the output gain of the RL was found to oscillate.
In Simulation 2, we confirmed the effectiveness of the proposed method in real environments with obstacles.
The results of the simulation experiment showed that the effectiveness of the controller using RL improved as the number of obstacles increased compared to the heuristic controller.
In addition, it was found that an RL controller trained in one real environment outperformed a heuristic controller when applied to other real environments.
This suggests that once an RL controller has been trained, it is sufficiently versatile to be used in different real environments.
In Simulation 3, we used multiple walking paths in VEs, and found that the number of resets decreased compared to when using the conventional general-purpose controller as the distance to the target increased.
These results suggest that the proposed method is effective when walking long distances in \red{real spaces} with obstacles.
In the user study, we confirmed the effectiveness of the proposed method in environments with/without obstacles.
The experimental results showed that the proposed controller outperforms the conventional controller.
Furthermore, the fact that there was no significant difference in VR sickness between the proposed method and the conventional method, it was suggests that the oscillation of the output gain did not cause VR sickness.

In future work, we aim to control multiple gains such as translation and rotation gains, which may improve the performance of the RL controller.
Furthermore, we will verify the effectiveness of the proposed method for dynamic obstacles and multiplayer situations where other players walk in the same real space at the same time.

\appendix
\section{\red{Hyperparameters}}
The hyperparameters were tuned manually as presented in Table \ref{table:hyperparameter}. 
Please check the ML-agents page\footnote {https://github.com/Unity-Technologies/ml-agents/blob/master/docs/Training-PPO.md} for details of each parameter.

\begin{table}[hbtp]
  \caption{Hyperparameter of PPO}
  \label{table:hyperparameter}
  \centering
  \begin{tabular}{lc}
    parameter & value \\
    \hline
    batch size &  2,048  \\
    buffer size  & 20,480  \\
    epsilon  & 0.2  \\
    gamma  & 0.995  \\
    number of layers & 2 \\
    hidden units & 128  \\
    lambda  & 0.995  \\
    learning rate  & $3 \times 10^{-4}$ \\
    max steps  & $1 \times 10^{6}$ \\
    memory size  & 256 \\
    num epoch  & 3 \\
    time horizon  & 256 \\
    sequence length  & 64 \\
    \hline
  \end{tabular}
\end{table}

\bibliographystyle{IEEEtran}
\bibliography{template}

\begin{IEEEbiography}[{\includegraphics[width=1in,height=1.25in,clip,keepaspectratio]{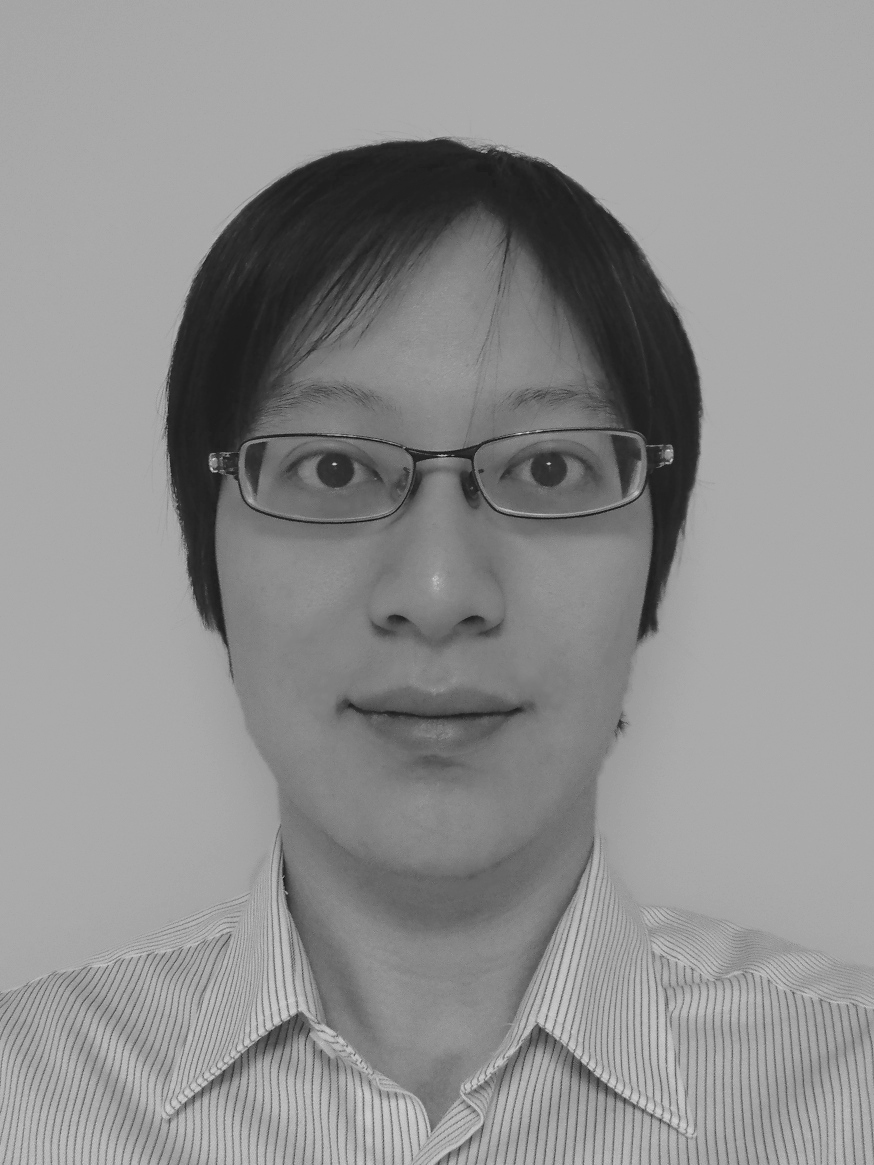}}]{Yuchen Chang} is a Game Engineer at GREE, Inc. He graduated from National Taiwan University with a bachelor's degree in electrical engineering and the University of Tokyo with a master's degree in mechano-informatics in 2020. His research interests include redirected walking and machine learning.
\end{IEEEbiography}

\begin{IEEEbiography}[{\includegraphics[width=1in,height=1.25in,clip,keepaspectratio]{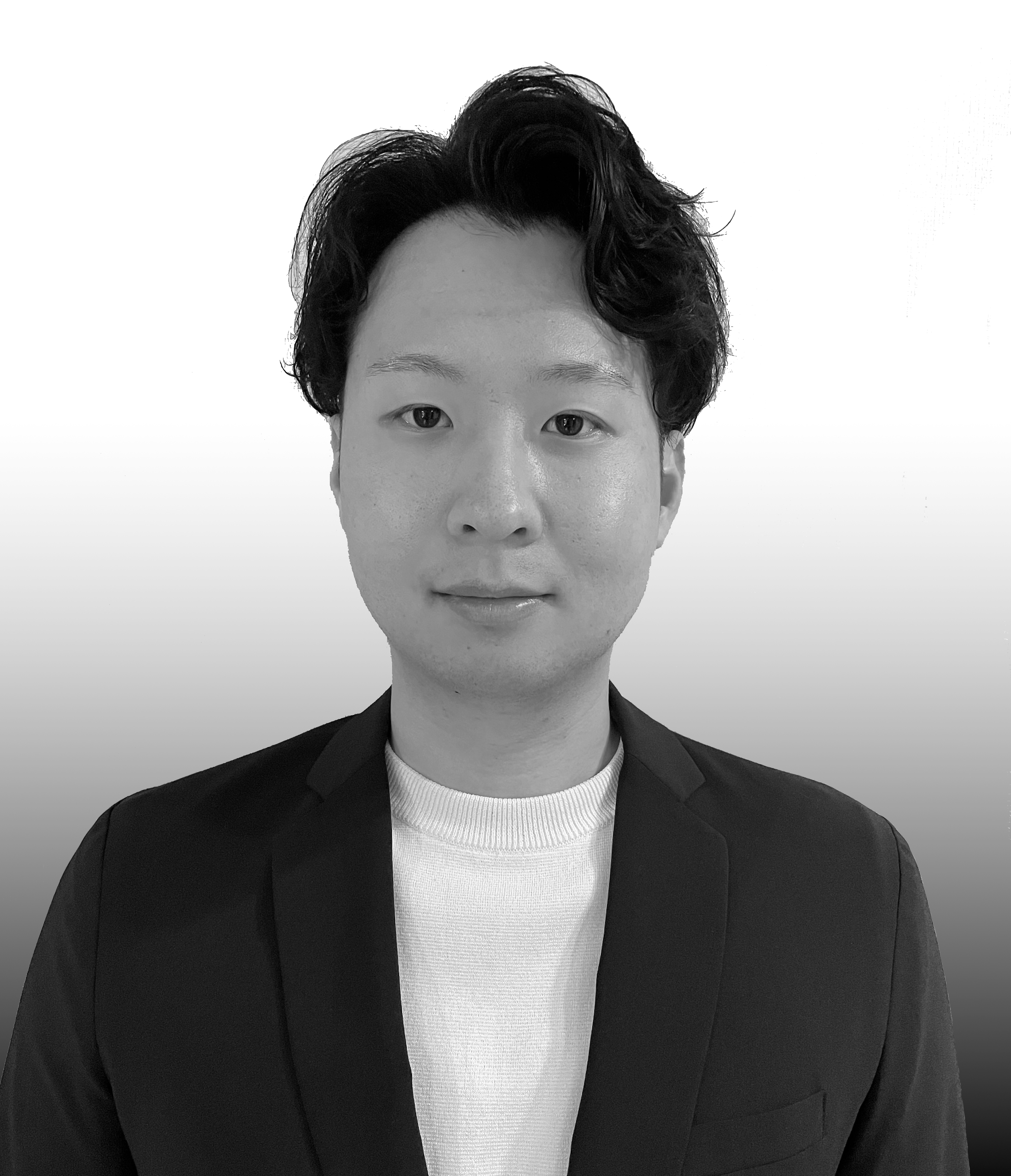}}]{Keigo Matsumoto}
is an Assistant Professor with the Graduate School of Information Science and Technology, the University of Tokyo. His research interests include spatial perception, redirected walking, and visuo-haptic interaction. He is a member of the ACM, IEEE, and the Virtual Reality Society of Japan (VRSJ). He received his B.E. and M.I.S.T degrees from the University of Tokyo in 2016 and 2018, respectively, and his Ph.D. degree in mechano-informatics from the University of Tokyo in 2021. He is the recipient of several awards, including Excellence Prizes in the 20th Japan Media Arts Festival Entertainment Division, SIGGRAPH ASIA 2017 E-tech Prize by the Award Committee, the President's award for Students of the University of Tokyo, and the VRSJ Outstanding Paper Award in the 24th Annual Meeting.
\end{IEEEbiography}

\begin{IEEEbiography}[{\includegraphics[width=1in,height=1.25in,clip,keepaspectratio]{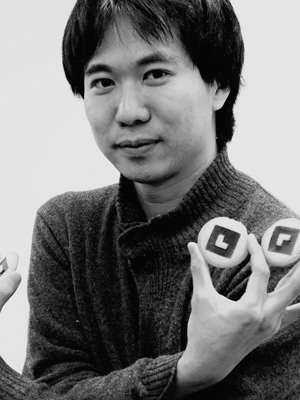}}]{Takuji Narumi}
is an Associate Professor with the Graduate School of Information Science and Technology, the University of Tokyo. His research interests broadly include perceptual modification and human augmentation with virtual reality and augmented reality technologies. He invented a novel haptic display, olfactory display, taste display, and satiety display by utilizing cross-modal interactions. More recently, he also invented an affective interface that which evokes specific emotions through pseudo-generated body reactions. He received his B.E. and M.E. degrees from the University of Tokyo in 2006 and 2008, respectively, and his Ph.D. degree in engineering from the University of Tokyo in 2011. He is the recipient of several awards including the ACE 2014 Gold Paper Award, ICAT 2012 Best Paper Award, Augmented Human 2016 Third Best Paper Award, ACM SUI2015 Short paper honorable mention, and VRSJ Outstanding Paper Award.
\end{IEEEbiography}

\begin{IEEEbiography}[{\includegraphics[width=1in,height=1.25in,clip,keepaspectratio]{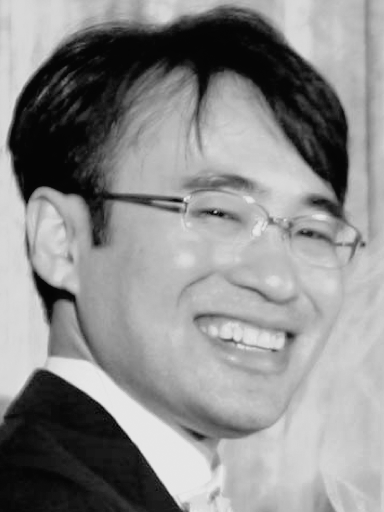}}]{Tomohiro Tanikawa}
is a Project Professor with the Graduate School of Information Science and Technology, the University of Tokyo. 
He received his M.Sc. and Ph.D. degrees in engineering from the Graduate School of Engineering, the University of Tokyo in 1998 and 2001, respectively, focusing on image-based 3D reconstruction and rendering methods. His research interests include image-based rendering, interactive computer graphics, virtual reality, and augmented reality. He has participated in several international and national research and development projects and has developed image-based rendering algorithms to generate high-quality images in real-time for the virtual environments and remote collaboration.
\end{IEEEbiography}

\begin{IEEEbiography}[{\includegraphics[width=1in,height=1.25in,clip,keepaspectratio]{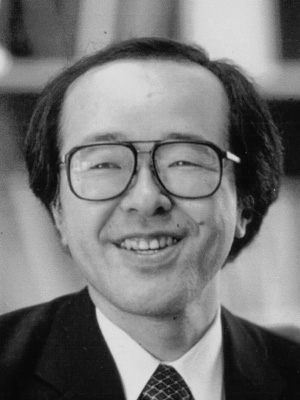}}]{Michitaka Hirose}
received his M.Sc. and Ph.D. degrees in engineering from the Graduate School of Engineering, the University of Tokyo in 1979 and 1982, respectively. His research interests include human–machine interaction, human-computer interface, interactive computer graphics, wearable computers, and virtual reality. He is in a position of leadership in the Japanese Virtual Reality Research community, and has been the Primary Investigator in several international and national research and development projects. He was a Project Leader of the Scalable Virtual Reality Project of the Telecommunications Advancement Organization (TAO) of Japan (the predecessor of NICT), and developed a tele-immersion communication system that connects between CAVE-type immersive environments through the large-bandwidth gigabit communication network. He is a member of the ACM and IEEE is a former president of the Virtual Reality Society of Japan (VRSJ).
\end{IEEEbiography}

\EOD

\end{document}